\documentclass[10pt,letter, review]{elsarticle}  %,review
\usepackage[letterpaper, total={6in, 8in}]{geometry}

\usepackage{lineno}
\modulolinenumbers[5]
\usepackage{amsmath,amssymb,amsmath,bm}
\usepackage[dvipsnames]{xcolor}  % {color}
\usepackage{times}
\usepackage{url,flushend,multirow,booktabs}
\usepackage{algorithm,algorithmic}
\usepackage{graphicx}
\usepackage{subfigure}
\usepackage{float}
\usepackage{hyphenat}
\usepackage[colorlinks,citecolor=green,urlcolor=blue,bookmarks=false,hypertexnames=true]{hyperref}
\usepackage{colortbl}
\definecolor{maroon}{cmyk}{0.08,0.04,0.00,0.06}  % light blue

\journal{arXiv}

\usepackage{numcompress}
\newcommand{\eg}{e.g.}
\newcommand{\ie}{i.e.}

\newcommand{\etal}{\textit{et al.}}
\newcommand{\aka}{\textit{a.k.a.}}

\renewcommand{\bm}{\mathbf}
\renewcommand{\b}{\color{blue}}

\DeclareFixedFont{\mf}{OT1}{ptm}{m}{n}{10pt}
\DeclareFixedFont{\mfb}{OT1}{ptm}{bx}{n}{10pt}

\begin{document}

\begin{frontmatter}

\title{Pseudo-labeling with Keyword Refining for Few-Supervised Video Captioning}

%% Include affiliations in footnotes:

%\ead[url]{www.elsevier.com}
\author[add1]{Ping~Li\corref{mark1}}
\cortext[mark1]{Corresponding author}
\ead{patriclouis.lee@gmail.com}
\author[add1]{Tao~Wang}
\author[add2]{Xinkui Zhao}
\author[add1]{Xianghua~Xu}
\author[add3]{Mingli~Song}

\address[add1]{School of Computer Science and Technology, Hangzhou Dianzi University, Hangzhou, China}
\address[add2]{School of Software Technology, Zhejiang University, Ningbo, China}
\address[add3]{College of Computer Science, Zhejiang University, Hangzhou, China}

\begin{abstract}
	Video captioning generate a sentence that describes the video content. Existing methods always require a number of captions (\eg, 10 or 20) per video to train the model, which is quite costly. In this work, we explore the possibility of using only one or very few ground-truth sentences, and introduce a new task named few-supervised video captioning. Specifically, we propose a few-supervised video captioning framework that consists of lexically constrained pseudo-labeling module and keyword-refined captioning module. Unlike the random sampling in natural language processing that may cause invalid modifications (\ie, edit words), the former module guides the model to edit words using some actions (\eg, copy, replace, insert, and delete) by a pretrained token-level classifier, and then fine-tunes candidate sentences by a pretrained language model. Meanwhile, the former employs the repetition penalized sampling to encourage the model to yield concise pseudo-labeled sentences with less repetition, and selects the most relevant sentences upon a pretrained video-text model. Moreover, to keep semantic consistency between pseudo-labeled sentences and video content, we develop the transformer-based keyword refiner with the video-keyword gated fusion strategy to emphasize more on relevant words. Extensive experiments on several benchmarks demonstrate the advantages of the proposed approach in both few-supervised and fully-supervised scenarios. The code implementation is available at https://github.com/mlvccn/PKG\_VidCap.   
\end{abstract}

\begin{keyword}
Video captioning \sep few supervision \sep pseudo-labeling \sep keyword refiner \sep gated fusion
\end{keyword}

\end{frontmatter}

%\linenumbers

\section{Introduction}
\label{sec1:intro}

Video captioning has established itself as a fundamental task in computer vision by describing video content using natural language sentences \cite{niu-pr2023-videocap}\cite{li-ipm2023-vidcap}. It is more complicated than image captioning \cite{cao-pr2024-imgcap} due to the existence of rich temporal relations and spatio-temporal dynamics in video. Generally, the generated captions are expected to accurately describe video content and adhere to the rules of grammar. It has found wide applications in many practical scenarios, such as video retrieval, the blind assistance, and human-computer interaction. However, existing methods are mostly fully-supervised methods, which require a large number of ground-truth descriptions (\eg, at least 20 or more for each video) to guide the model training, which is very costly due to expensive human labeling. This raises a problem that how to guarantee the captioning quality when only single or very few (\eg, less than three) ground-truth descriptions (\aka, sentences) are available. Little effort has been made to consider this setting, which is named as \emph{few-supervised video captioning} in this work. Note that it is different from few-shot setting  (only a few samples per class) or semi-supervised setting (using both labeled and unlabeled data) in machine learning.

Existing fully-supervised methods always adopt the encoder-decoder framework, \eg, encoder uses ResNet (Residual Network) \cite{he-cvpr2016-resnet}, Faster R-CNN \cite{ren-tpami2017-faster-rcnn} (Regional Convolutional Neural Network), and C3D (Convolutional 3D network) \cite{tran-iccv2015-c3d} to extract appearance, object, and motion features respectively, while decoder employs LSTM (Long-Short Term Memory) to decode video features into a descriptive sentence. For example, Gao \etal~\cite{gao-tip2022-hrnat} employ hierarchical attention mechanism as well as the knowledge and parameter sharing in language generation module to obtain hierarchical representations with rich semantics. Recently, transformer has received much interests in video captioning, \eg, Ye \etal~\cite{ye-cvpr2022-hmn} extract entity-level features by transformer to capture the most possibly referred object in caption. However, all these methods require an abundant of complete video-text pairs during training, which consumes very expensive labeling costs. Unlike them or weakly supervised learning which employs video-level annotation, there is only single or very few ground-truth sentence for each video in few-supervised setting. By contrast, few-supervised video captioning aims to approach or achieve the comparable performance of fully-supervised scenario. 

% -------------------------  Motivation for few-supervised video captioning -------------------
\begin{figure}[!t]
	\centering
	\includegraphics[width=\linewidth]{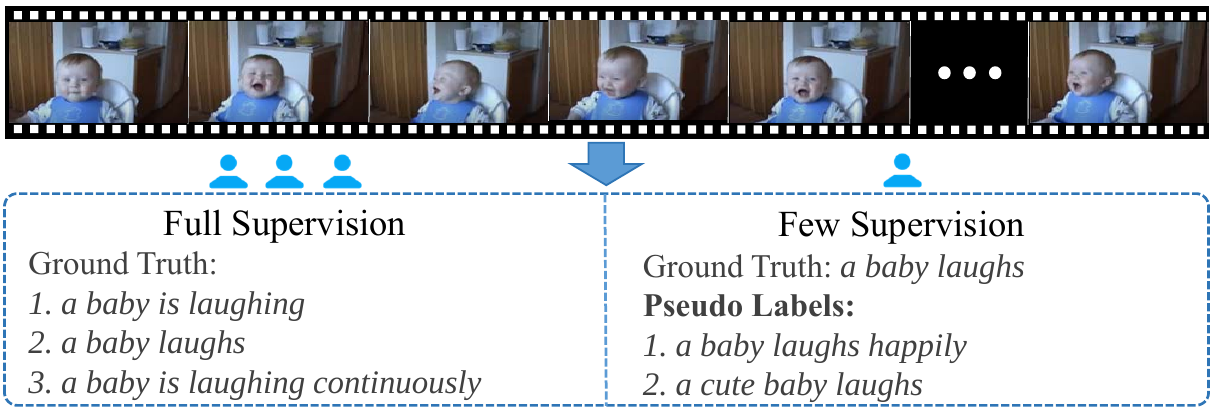}
	\caption{Motivation illustration (3 vs 1 human). Video is from MSVD \cite{chen-acl2011-msvd}.}
	\label{fig:motivation}
	\vspace{-3mm}
\end{figure}

As depicted in Fig.~\ref{fig:motivation}, full supervision needs three human annotations while few-supervision only needs one human annotation, largely saving the labeling costs. To bridge the semantic gap between full supervision and few supervision, there are \textbf{two} primary challenges, \ie, 1) \emph{how to expand sentences by pseudo labeling}, and 2) \emph{how to guarantee the high-quality pseudo-labeled sentences for model training}.

Naturally, we may resort to data augmentation for expanding sentences by adopting the Natural Language Processing (NLP) techniques, such as synonym replacement and replacement with grammar rules. But they neglect sentence context which may cause the ambiguous problem, because a word has multiple synonyms which represent distinct meanings in a sentence. For example, \emph{``A dog is giving a \textbf{high} five to man''} is edited as \emph{``A dog is giving a \textbf{tall} five to man''}, which changes the original meaning of \emph{``high five''}. Besides, one can use a deep learning language model \cite{hou-aaai2021-c2c-genda} to randomly replace a word, or translate between two languages, which considers the context but hard to keep still the semantics of some key words (\ie, nouns and verbs). To address this issue, we employ a lexical constrained sentence generation technique that makes sentence include specified words or phrases. In particular, we adopt a two-step strategy \cite{he-aaai2021-mcmc} with lexical constraints, by first employing pretrained word classifier to guide the model to edit words using some action (\eg, copy, replace, insert, and delete) and then fine-tuning generated sentences by pretrained language model (\eg, XLNet~\cite{yang-nips2019-xlnet}), leading to \emph{candidate pseudo-labeled sentences}. Moreover, we use the repetitive penalized sampling to impose the penalty when repeated words appear, which reduces repetitions in sentence. 

Having obtained candidate sentences, how to effectively employ them for model training poses a major challenge. The above data augmentation neglects visual cues, which may cause inconsistency between pseudo-labeled sentences and video content, leading to noisy pseudo labels. To overcome this drawback, we introduce a pretrained video-text model, \ie, X-CLIP~\cite{ni-eccv2022-xclip}, for matching the most relevant candidate sentences with video, \ie, \emph{pseudo-labeled sentences}. These pseudo-labeled sentences and human-labeled sentence are both fed into the captioning model as augmented supervision knowledge to guide the model training. To enhance the captioning ability of model, we design a video-keyword gated fusion scheme in the captioning model, which employs the attention mechanism in transformer to adjust the weights of key words in pseudo-labeled sentences. In particular, it increases the weights of the key words more relevant to video while decreasing the weights of those less relevant or irrelevant to video. As expected, the irrelevant words are overlooked and the relevant ones are more emphasized in pseudo-labeled sentence. In addition, we design a semantic loss of key words between pseudo-labeled and human-labeled sentences to ensure their semantic consistency, which is beneficial for improving the generalization ability of the model yielding both more fluent and more accurate sentences for a given video.

Therefore, we propose a \textbf{P}seudo-labeling with \textbf{K}eyword-refiner and \textbf{G}ated fusion (\textbf{PKG}) approach for few-supervised video captioning. It consists of two primary components, \ie, lexically constrained pseudo-labeling module and keyword-refined captioning module. The former is designed to generate pseudo-labeled sentences to compensate for the lack of sufficient supervision knowledge during model training, while the latter aims to match those most relevant candidate sentences with video by modeling the global context of visual cues and language semantics. To investigate the performance of our method, a large number of experiments and ablation studies were carried out on three publicly available data sets including MSVD~\cite{chen-acl2011-msvd}, MSR-VTT~\cite{xu-cvpr2016-msr-vtt}, and VATEX~\cite{wang-iccv2019-vatex}. 

The main contributions of this paper are summarized below:
\begin{itemize}
	\item We introduce a new task named few-supervised video captioning, which uses only one human-labeled (ground-truth) sentence to train the model.
	\item We propose a pseudo labeling strategy with lexical constraint to augment supervision knowledge, which strengthens the guidance of model training.
	\item We design a keyword-refined captioning module with video-text gated fusion for generating high-quality sentences that matches video content by modeling the global context. 
	\item Empirical studies on several benchmarks demonstrate our approach using only one human-labeled sentence achieves more promising performance than those using more human-labeled sentences. In addition, our gated fusion captioning model surpasses State-Of-The-Art (SOTA) supervised methods when all ground-truth labels are used.
	
\end{itemize}
%

%-------------------------------------------------------------------------
\section{Related Work}
\label{related}
This section mainly discusses several closely related works including video captioning and data augmentation in Natural Language Processing (NLP). 

\subsection{Video Captioning}
Generally video captioning aims to describe video content using one sentence, which differs from dense video captioning which describe complex video via multiple sentences. Traditional video captioning methods mainly generate sentences according to the template that adheres to some grammar rules, but they heavily rely on template which usually leads to fixed sentence form without flexibility and diversity.  

Recently, the success of deep learning in image processing has expedited its application to video captioning \cite{li-neucom2022-gcn} by adopting the encoder-decoder framework, in which encoder learns visual features and decoder yields sentences. The earlier work \cite{venugopalan-iccv2015-s2vt} proposes a sequence to sequence method that employs Long Short-Term Memory (LSTM) to capture temporal relations, and later bidirectional LSTM and hierarchical LSTM are used to better capture temporal cues. As known to us, the motion information plays an important role in modeling dynamics in video, \eg, Chen~\etal~\cite{chen-iccv2021-mgrmp} employ motion-guided region message passing scheme to extract object spatial features. Besides, object relation benefits generating high-quality sentences, and there are several works concentrating on exploring object feature learning and object relation modeling. For example, from the multi-level feature perspective, Dong~\etal~\cite{dong-acmmm2021-multi} utilize multi-head self-attention mechanism to enhance the scene-level and object-level features separately before decoding; from the object-level interaction perspective, Bai~\etal~\cite{bai-acmmm2021-dlsg} build a discriminative latent semantic graph to fuse spatio-temporal information into latent object proposal for generating semantic-rich captions. However, these methods depend heavily on pre-trained detectors of objects and their relationships. To address this issue, Jing~\etal~\cite{jing-tmm2023-man} build a memory-based augmentation network to incorporate implicit external knowledge with a neural memory.
 
Furthermore, attention-based methods are extensively used for generating captions. They identify those most relevant frame features by adaptive weighting, which enhances the semantic alignment between video frames and their corresponding words. For example,  Yang~\etal~\cite{yang-aaai2021-nacf} employ bidirectional self-attention to capture more visual details in a non-autoregressive coarse-to-fine framework; Verma~\etal~\cite{verma-aaai2023-vsvcap} exploit a fine-grained cross-graph attention mechanism to capture detailed graph embedding correspondence between visual graph and textual knowledge graph, narrowing down the visual-text gap; Gao~\etal~\cite{gao-tip2022-hrnat} attempt to learn multi-level representations (\ie, objects, actions and events) by hierarchical attention and cross-modality matching; Tu~\etal~\cite{tu-pr2023-videocap} present a relation-aware attention mechanism with graph learning for video captioning; Luo~\etal~\cite{luo-pr2024-videocap} design a feature aggregation module based on a lightweight attention mechanism to aggregate frame-level features. However, these methods adopt some backbone to obtain video features at first which are unnecessarily suitable for video captioning. To handle this shortcoming, Lin~\etal~\cite{lin-cvpr2022-swinbert} propose a transformer-based end-to-end video captioning model with sparse attention. Unlike the above works, we design a video-keyword gated fusion scheme that employs the attention mechanism in transformer to adjust the weights of key words in pseudo-labeled sentences. This allows to increase the weights of the key words more relevant to video while decreasing the weights of those less relevant or irrelevant to video.

In addition, there are some works dealing with the long-tail issue of words, which cause the model to neglect low-frequency but important words (\aka~tokens). For example, Gu~\etal~\cite{gu-cvpr2023-textkg} present a text with knowledge graph augmented transformer which includes an external stream and an internal stream, where the latter exploits the multi-modality information (\eg, frame appearance, speech transcripts, and captions) to alleviate the long-tail problem. Nevertheless, all the above methods desire full supervision, and fail to handle the scenario where only very few supervision knowledge is available for model training.

{\b In some sense, ``few-supervised'' seems to be related to ``few-shot video captioning'' \cite{Alayrac-neurips2022-flamingo, Wang-neurips2022-vidil} , however, they are totally different. we use the word ``few-shot'' to indicate the scenario where only a single or very few (say less than three) sentences are labeled for a video, \ie, the supervised knowledge is very limited. By contrary, ``few-shot captioning'' indicates that there are very few prompt instances and their models are build upon the large model with carefully-designed prompts, such as descriptions about object and actions; these models often employ a few instances as prompts for the well-trained Large Language Model (LLM) to generate captions, and they neither train the model nor change the architecture.}

\subsection{Data Augmentation in NLP} 
The most popular data augmentation technique is paraphrasing in NLP, which rephrases or restates a text segment using different words and sentence structure. Generally, it contains three generation modes, \ie, word-level, phrase-level, and sentence-level. For \emph{word-level} mode, it is common to use synonym substitution by virtue of synonym dictionary such as WordNet, and adopts random substitution to replace multiple words, which neglects word context. This may lead to the word ambiguous problem, which inspires Wu~\etal~\cite{wu-iccs2019-conditional} to fine-tune from the pretrained language representation model BERT (Bidirectional Encoder Representations from Transformers) such that sentences are augmented. For \emph{phrase-level} mode, Wang \etal~\cite{wang-emnlp2015-that} employ word embedding and phrase embedding; for the former the source word is replaced by nearest-neighbour word in terms of cosine similarity; for the latter the Word2Vec is used to extract phrase embedding and replace the current phrase by its most relevant ones. For \emph{sentence-level} mode, Hou \etal~\cite{hou-aaai2021-c2c-genda} employ a multi-layer transformer to encode multiple neighboring sentences, and harness a duplication-aware attention scheme to generate distinct variations of sentences.

Although the aforementioned methods are capable of generating sentences with certain semantic similarity, they overlook sentence context when adopting random synonym substitution or semantic embedding replacement. This oversight leads to word or phrase ambiguous issues, \ie, different synonyms can carry distinct meanings within the same sentence context. Moreover, they just rely on substitution technique and fail to add or remove words from sentences, thus limiting the diversity of sentence structures. To overcome these shortcomings, we present a lexically constrained pseudo-labeling approach that promotes generating diverse and less repetitive captions in few-supervised scenario.
%
% -------------------------  Framework of few-supervised video captioning -------------------
\begin{figure*}[!t]
	\centering
	\includegraphics[width=0.92\linewidth]{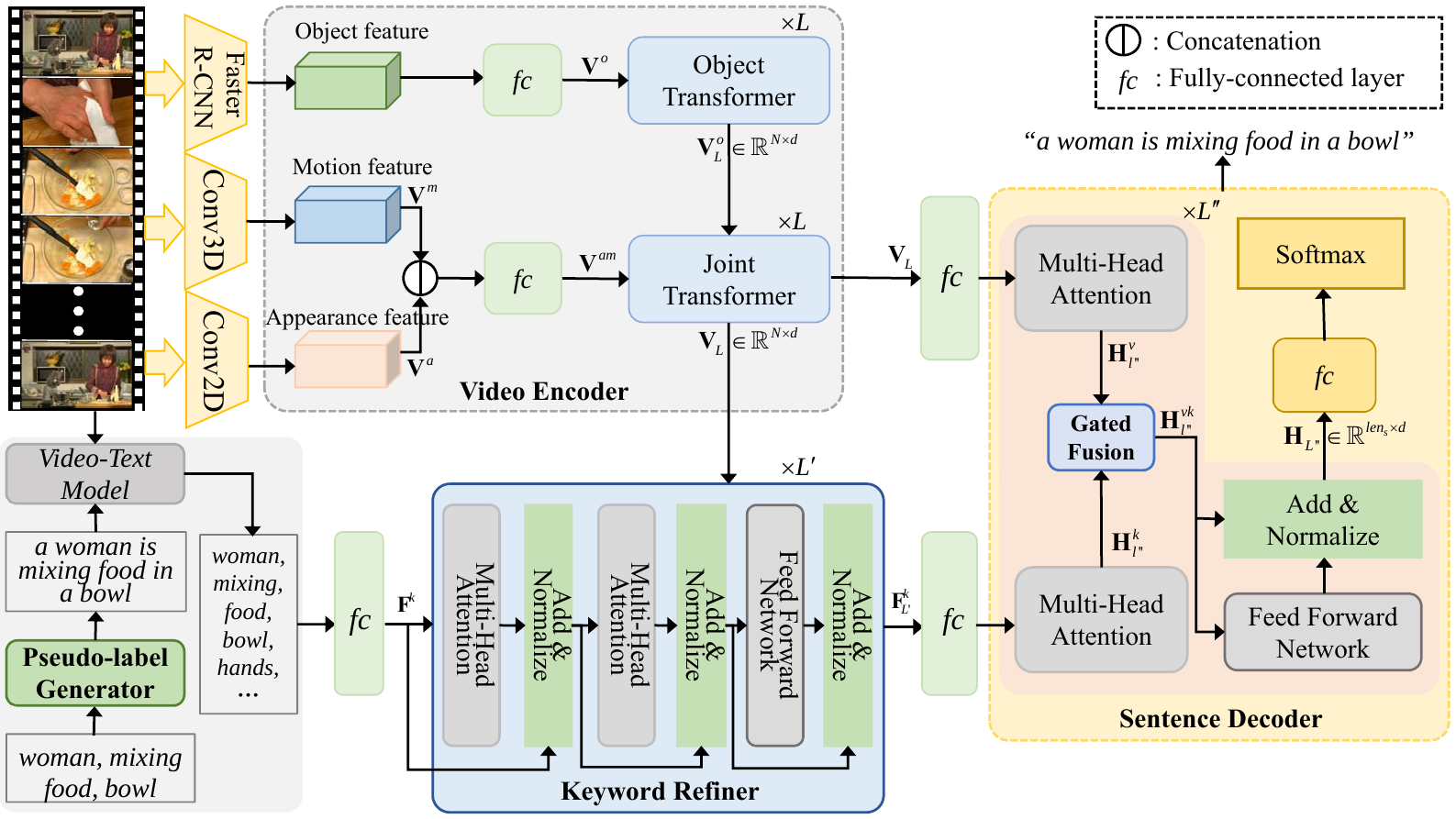}
	\caption{Overall framework of Pseudo-labeling with Keyword-refiner and Gated fusion (PKG) method for few-supervised video captioning.}
	\label{fig:framework}
	%\vspace{-3mm}
\end{figure*}

%------------- Methodology ---------
\section{Method}
\label{method}
\begin{figure}[!t]
	\centering
	\includegraphics[width=\linewidth]{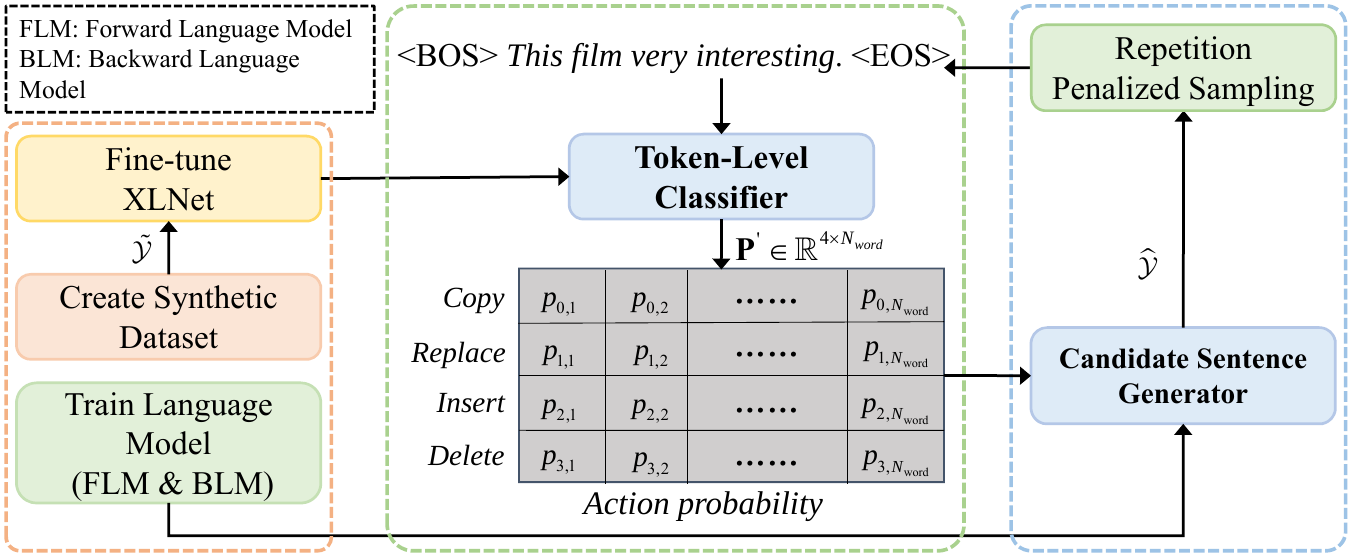}
	\caption{Pseudo-label generator.}
	\label{fig:pseudolabeling}
	%\vspace{-3mm}
\end{figure}

This section presents our proposed Pseudo-labeling with Keyword-refiner and Gated fusion (PKG) method for few-supervised video captioning. The overall framework is illustrated in Fig.~\ref{fig:framework}, which has two primary components, \ie, the \textit{Pseudo-labeling module} (left bottom, see details in Fig.~\ref{fig:pseudolabeling}), including synthetic data creator, token-level classifier, and candidate sentence generator, and the \textit{Captioning module} including video encoder, keyword refiner, and sentence decoder. Details are described below.

% -------------------------  Framework of pseudo-labeling module -------------------

\subsection{Problem Definition}
Few-supervised video captioning aims to generate natural language descriptions about video content by using only one or very few ground-truth sentences. Formally, given a video $\mathcal{X} = \{\mathbf{X}_i \in \mathbb{R}^{w \times h \times 3} | 1 \le i \le N_f \}$ that consists of $N_f$ frames, where $\bm{X}_i$ denotes the $i$-th frame with the width $w$, the height $h$, and RGB three channels, it has a set of $N_{gt}$ ground-truth sentences $\Phi = \{\mathcal{Y}_1, \mathcal{Y}_2, ..., \mathcal{Y}_{N_{gt}} \}$. In few-supervised setting, $N_{gt}$ usually takes 1 or 2 which is much less than that used in fully-supervised setting, and the ground-truth sentence is $\mathcal{Y} = \{ \mathbf{y}_t \in \mathbb{R}^{N_{voc}}| 1 \le t \le len_s\}$, where $\bm{y}_t$ is the one-hot vector of the $t$-th word, $len_s$ denotes sentence length, and $N_{voc}$ is the vocabulary size (\aka, total number of words). Note that the words in vocabulary are those appearing at least twice in ground-truth sentence set. We use the \emph{Pseudo-labeling module} to generate pseudo-labeled sentence $\hat{\mathcal{Y}} = \{\hat{\mathbf{y}}_t \in \mathbb{R}^{N_{voc}}| 1 \le t \le len_s\}$ by utilizing the keyword sequence $\mathcal{Y}^k = \{ \mathbf{y}_{t'}^k \in \mathbb{R}^{N_{voc}} | 1 \le t' \le N_{word}\}$, where $\bm{y}_{t'}^k$ is the one-hot vector of the $t'$-th keyword, and the superscript ‘k’ denotes keyword. Here, keywords  involving nouns and verbs are extracted from ground-truth sentence $\mathcal{Y}$. Then, we employ the \emph{Captioning module} to yield predicted sentence $\mathcal{Y}^\prime = \{\mathbf{y}_t^\prime \in \mathbb{R}^{N_{voc}}| 1 \le t \le len_s\}$. 

\subsection{Pseudo-labeling Module}
The pseudo-labeling module mainly consists of synthetic data creator, token-level classifier, and candidate sentence generator. Here we adopt the lexically-constrained data augmentation to generate pseudo-labeled sentence, and details are shown below. 

\textbf{Synthetic Data Creator}. As mentioned earlier, there are four actions to edit words in sentence, \ie, copy, replace, insert, and delete. To represent these actions for the words, as in \cite{he-aaai2021-mcmc}, we create a synthetic data set $\mathcal{S}$ which is composed of many data pairs, \ie, $(\tilde{\mathcal{Y}}, \mathcal{C})$, where $\tilde{\mathcal{Y}} = \{\tilde{\bm{y}}_{t''}\in \mathbb{R}^{N_{voc}} | 1\le {t''} \le len_{\tilde{s}} \}$ is a synthetic sentence, $\tilde{\bm{y}}_{t''}$ is the one-hot vector of the $t''$-th token (word), $len_{\tilde{s}}$ is the length of synthetic sentence, and $\mathcal{C} = \{ c_{t''} \in \{0, 1, 2, 3\} | 1 \le {t''} \le len_{\tilde{s}} \}$ represent the action of each token, \ie, $\{0, 1, 2, 3\}$ denote \emph{copy}, \emph{replace}, \emph{insert}, and \emph{delete}, respectively. Note that word and token are the same meaning, which are interchangeably used here. Generally, we create the synthetic data in the following way. 

Given a sentence ``\emph{A woman and her dog walking down a side walk next to a fence with some flowers.} ", we truncate its segment to obtain ``\emph{A woman and her dog walking down}", and create different synthetic pairs for different actions. For instance, given $\tilde{\mathcal{Y}} = \{BOS, woman, her, dog, walking, down, EOS\}$, where $BOS$ is sentence start and $EOS$ is sentence end, we do the ``\emph{insert}'' action according to $\mathcal{C} = \{0,2,2,0,0,0,2\}$, where the 2nd, the 3rd, and the last elements take the value 2. This means it requires inserting some word ``A'', ``and'', and ``a'' before \emph{her}, \emph{woman}, and $EOS$. Note that we adopt the masked language model XLNet \cite{yang-nips2019-xlnet} to create synthetic pairs for ``\emph{replace}'' and ``\emph{delete}'' actions. Given $\tilde{\mathcal{Y}} = \{BOS, A, woman, and, her, MASK, dog, \\ walking, down, a, EOS\}$, we do the ``\emph{delete}'' and ``\emph{insert}'' actions according to $\mathcal{C} = \{0,0,0,0,0,3,0,0,0,0,2\}$. Then we use XLNet to predict the $MASK$ (the 6-th element) by selecting the token ``small" with the highest probability, leading to the final synthetic sentence $\tilde{\mathcal{Y}} = \{BOS, A, woman, and, her, small, dog, walking, down, a, \\ EOS\}$. Here, the token ``small"  (the 6-th element with value 3) should be deleted, and before token $EOS$ we should insert the token ``sidewalk". Given $\tilde{\mathcal{Y}} = \{BOS, A, woman, and, her, MASK, walking, down, a, EOS\}$, we do the ``\emph{replace}'' action according to $\mathcal{C} = \{0,0,0,0,0,1,0,0,0,2\}$, where the token \emph{MASK}  (the 6-th element with value `1' should be replaced with the token \emph{baby} with the highest probability predicted by XLNet, leading to the final synthetic sentence $\tilde{\mathcal{Y}} = \{BOS, A, woman, and, her, baby, walking, down, a, EOS\}$.

\textbf{Token-level Classifier}. After obtaining the synthetic data pairs $(\tilde{\mathcal{Y}}, \mathcal{C})$, we finetune the language model XLNet \cite{yang-nips2019-xlnet} on them, which results in a token-level classifier. During finetuning, we predict the action by XLNet for the tokens in synthetic sentences. Then we employ this token-level classifier to compute the prior probability of different actions for the tokens. In detail, we use text analysis tool \footnote{nltk: \url{https://www.nltk.org/}} to extract key words from ground-truth sentence, including nouns and verbs, for constituting a keyword sequence, \ie, $\mathcal{Y}^k = \{ \mathbf{y}_{t'}^k \in \mathbb{R}^{N_{voc}} | 1 \le t' \le N_{word}\}$, where $\bm{y}_{t'}^k$ is the one-hot vector of the $t'$-th one from the total $N_{word}$ key words. For example, given a sentence ``\emph{A man is playing guitar}", we first obtain its keyword sequence $\mathcal{Y}^k = \{BOS, man, playing, guitar, EOS \}$, and then send such sequence to the token-level classifier. This yields an action probability matrix $\mathbf{P}^\prime \in \mathbb{R}^{4 \times N_{word}}$, whose entry $p'_{r,t'}$ denotes the probability of the $r$-th action for the $t'$-th token in sentence, and $p'_r$ denotes the probability of the $r$-th action by summing up the row entries. We choose the action with the highest probability as the current action, and then choose the token with the highest probability as the current token to be edited. 

\textbf{Candidate Sentence Generator}. Given the current candidate pseudo-labeled sentence $\hat{\mathcal{Y}} = \{ \hat{\mathbf{y}}_t \in \mathbb{R}^{N_{voc}} | 1 \le t \le len_s\}$, where $\hat{\mathbf{y}}_t$ denotes the $t$-th token, we assume the current action is ``\emph{replace}" and the position is the $t$-th token. Now we need to compute the probability of the $t$-th token to be replaced. In particular, we pretrain the language model XLNet \cite{yang-nips2019-xlnet} , \ie, Forward Language Model (FLM) and Backward Language Model (BLM), using the image captions of MSCOCO database \cite{lin-eccv2014-mscoco}. To predict the next token, the former FLM inputs the captions to XLNet in a forward manner while the latter BLM acts in a reverse manner. To replace the $t$-th token, FLM takes the token sequence before $\hat{\mathbf{y}}_t$ as input, \ie, $\hat{\mathcal{Y}}_{1 \rightarrow (t-1)} = \{\hat{\mathbf{y}}_1, \cdots, \hat{\mathbf{y}}_{t-1} \}$, and BLM takes that after $\hat{\mathbf{y}}_t$ as input in a reverse order, \ie, $\hat{\mathcal{Y}}_{len_s \rightarrow (t+1)} = \{\hat{\mathbf{y}}_{len_s}, \cdots, \hat{\mathbf{y}}_{t+1} \}$. 

Mathematically, we calculate the probability of replacing the $t$-th token using FLM $\hat{p}^{FLM} = p^{FLM}(\hat{\mathbf{y}}_t = \mathbf{y}_j^{voc} | \hat{\mathcal{Y}}_{1 \rightarrow (t-1)})$ and BLM $\hat{p}^{BLM} = p^{BLM}(\hat{\mathbf{y}}_t = \mathbf{y}_j^{voc} | \hat{\mathcal{Y}}_{len_s \rightarrow (t+1)}) $ respectively by
\begin{equation}
	\label{eq:flm}
	\hat{p}^{FLM} = \frac{p^{FLM}(\hat{\mathcal{Y}}_{1 \rightarrow (t-1)} | \hat{\mathbf{y}}_t=\mathbf{y}_j^{voc}) / I(\mathbf{y}_j^{voc} \in \hat{\mathcal{Y}})}
	{\sum_{j=1}^{N_{voc}} p^{FLM}(\hat{\mathcal{Y}}_{1 \rightarrow (t-1)} | \hat{\mathbf{y}}_t=\mathbf{y}_j^{voc} ) / I(\mathbf{y}_j^{voc} \in \hat{\mathcal{Y}})},
\end{equation}

\begin{equation}
	\label{eq:blm}
	\hat{p}^{BLM}= \frac{p^{FLM}(\hat{\mathcal{Y}}_{len_s \rightarrow (t+1)} | \hat{\mathbf{y}}_t=\mathbf{y}_j^{voc}) / I(\mathbf{y}_j^{voc} \in \hat{\mathcal{Y}})}
	{\sum_{j=1}^{N_{voc}} p^{FLM}(\hat{\mathcal{Y}}_{len_s \rightarrow (t+1)}| \hat{\mathbf{y}}_t=\mathbf{y}_j^{voc} ) / I(\mathbf{y}_j^{voc} \in \hat{\mathcal{Y}})},
\end{equation}
where $\mathbf{y}_j^{voc}\in \mathbb{R}^{N_{voc}}$ denotes the $j$-th token in vocabulary, . Here, we adopt the repetition penalized sampling strategy, which discounts the scores of previously generated tokens by the penalty function $I(\cdot)$. Empirically, this function takes 1.2 or 1 when a word appears or does not appear in a sentence. In another word, when a word has  already appeared in a sentence, we lower down its likelihood of being included again, and vice versa. This alleviates the word repetition problem to some degree. By multiplying of the above two probabilities in Eq.~(\ref{eq:flm}) and Eq.~(\ref{eq:blm}), we obtain the probability of replacing the $t$-th token, \ie, $\hat{p} = \hat{p}^{FLM}\cdot \hat{p}^{BLM}$, where we omit the superscript $t$ for brevity. 

After calculating all tokens in vocabulary, we choose the token with the maximum probability to replace the current one. For the ``\emph{insert}" action, we perform the similar procedures. Note that we simply copy or delete the word for the ``\emph{copy}" or ``\emph{delete}" action. When we repeat $T$ (empirically set to 10) times, we obtain $T$ candidate pseudo-labeled sentences. However, these sentences do unnecessarily cater to video content, so we attempt to identify those most relevant ones by adopting a pretrained video-text model X-CLIP \cite{ni-eccv2022-xclip}, for extracting video and text features. Specifically, we compute the cosine similarity of the video feature $\bm{f}_v \in \mathbb{R}^d$ and the text feature $\bm{f}_t \in \mathbb{R}^d$ of candidate pseudo-labeled sentence. Then, we list the similarity scores of all sentences in a descending order, and select the leading $N_{pse}$ ones to form the pseudo-labeled sentence set $\hat{\Phi} = \{ \hat{\mathcal{Y}}_1, \hat{\mathcal{Y}}_2, ..., \hat{\mathcal{Y}}_{N_{pse}} \}$.

\subsection{Captioning Module}
The captioning module mainly consists of video encoder, keyword refiner, and sentence decoder. Details are shown below.

\textbf{Video Encoder}. We adopt the transformer \cite{vaswani-nips2017-transformer} block to build video encoder to learn global features. First, we obtain the appearance feature set $\mathcal{V}^a = \{ \mathbf{v}_{i}^a \in \mathbb{R}^{d_a} | 1 \le i \le N_f\}$ by Inception ResNetv2 \cite{szegedy-aaai2017-inception-v4}, and the motion feature set $\mathcal{V}^m = \{ \mathbf{v}_{\hat{i}}^m \in \mathbb{R}^{d_m} | 1 \le \hat{i} \le N_{clip} \}$ by C3D (Convolutional 3D) \cite{hara-cvpr2018-c3d}, where the superscript ``a'' denotes appearance, ``m'' denotes motion, and $\hat{i}$ indexes the 16-frame clip in a video. From both of them, we evenly select $N$ feature vectors and reshape them to an appearance matrix $\mathbf{V}^a \in \mathbb{R}^{N \times d_a}$ and an motion matrix $\mathbf{V}^m \in \mathbb{R}^{N \times d_m}$, which are concatenated into an appearance-motion feature matrix $\hat{\mathbf{V}}^{am} \in \mathbb{R}^{N \times (d_a+d_m)}$. Then, we obtain the object feature set $\mathcal{V}^o = \{\mathbf{v}_z^o \in \mathbb{R}^{d_o} | 1 \le z \le N',\}$ by Faster R-CNN \cite{ren-tpami2017-faster-rcnn}, where the superscript ``o'' denotes object, $z$ is object index, and $N'=10\cdot N_{clip}$ is total object number (each clip has 10 objects). From the set, we evenly select $N_{obj}$ feature vectors and reshape them to an object feature matrix $\hat{\mathbf{V}}^o \in \mathbb{R}^{N_{obj} \times d_o}$. To keep the same feature dimension $d$, we apply a fully-connected (fc) layer to the appearance-motion features and the object features, resulting in $\mathbf{V}^{am} \in \mathbb{R}^{N \times d}$ and $\mathbf{V}^o \in \mathbb{R}^{N_{obj} \times d}$, respectively. 

For object features, it desires to select those important objects relevant to captions. Hence, we design an object transformer that consists of $L$ transformer blocks, and each block is composed of one multi-head attention layer, twice layer normalizations, and one feed-forward network. The output $\bm{V}_{l-1}^o\in \mathbb{R}^{N_{obj} \times d}$ of the previous ($l$-1)-th block is the input of the current $l$-th block, and the input of the first block is $\mathbf{V}^o$. Formally, the output $\bm{V}_l^o$ of the $l$-th block is obtained by
\begin{equation}
	\begin{aligned}
		\label{eq:obj_fea_output}
		\hat{\mathbf{V}}_l^o & = \mathrm{LayerNorm}(\mathbf{V}_{l-1}^o + \mathrm{MultiHead}(\mathbf{V}_{l-1}^o, \mathbf{V}_{l-1}^o, \mathbf{V}_{l-1}^o)), \\
		\mathbf{V}_{l}^o & = \mathrm{LayerNorm}(\hat{\mathbf{V}}_l^o + \mathrm{FFN}(\hat{\mathbf{V}}_l^o)),
	\end{aligned}
\end{equation}
where $\mathrm{LayerNorm}(\cdot)$ is the layer normalization, $\mathrm{FFN}(\cdot)$ is a two-layer feed-forward network, and $\mathrm{MultiHead}(\cdot)$ is multi-head attention which accepts \emph{query}, \emph{key}, and \emph{value} as its input. As we see, the temporary feature $\hat{\mathbf{V}}_l^o$ is obtained by performing the self-attention on $\mathbf{V}_{l-1}^o$ with residual connection and layer normalization. Therefore, the final output of the object transformer is $\bm{V}_L^o\in \mathbb{R}^{N_{obj} \times d}$. 

To consider object features together with appearance-motion features, we design a joint transformer that also consists of $L$ transformer blocks, and each block is composed of two multi-head attention layers, three layer normalizations, and one feed-forward network. The output $\bm{V}_{l-1} \in \mathbb{R}^{N\times d}$ of the previous ($l$-1)-th block is the input of the current $l$-th block, and the inputs of the first block are the final object feature $\mathbf{V}_L^o$ and the appearance-motion feature $\mathbf{V}^{am}$. Formally, the output $\bm{V}_l^o$ of the $l$-th block is obtained by 
\begin{equation}
	\begin{aligned}
		\label{eq:joint_fea_output}
		\hat{\mathbf{V}}_l & = \mathrm{LayerNorm}(\mathbf{V}_{l-1} + \mathrm{MultiHead}(\mathbf{V}_{l-1}, \mathbf{V}_{l-1}, \mathbf{V}_{l-1})), \\
		\tilde{\mathbf{V}}_l & = \mathrm{LayerNorm}(\hat{\mathbf{V}}_l + \mathrm{MultiHead}(\hat{\mathbf{V}}_l, \bm{V}_L^o, \bm{V}_L^o)), \\
		\mathbf{V}_{l} & = \mathrm{LayerNorm}(\tilde{\mathbf{V}}_l + \mathrm{FFN}(\tilde{\mathbf{V}}_l)),
	\end{aligned}
\end{equation}
where the temporary feature $\hat{\mathbf{V}}_l$ is obtained by performing the self-attention on $\mathbf{V}_{l-1}$ with residual connection and layer normalization, and the temporary feature $\tilde{\mathbf{V}}_l$ is obtained by performing the cross attention on appearance-motion feature $\mathbf{V}_{l-1}$ and object feature $\bm{V}_L^o$. Therefore, the final output of the joint transformer is video feature $\bm{V}_L \in \mathbb{R}^{N \times d}$. 

\textbf{Keyword Refiner}. When generating pseudo-labeled sentences, some keywords may not adhere to video content. Moreover, the keywords in both pseudo-labeled and ground-truth sentences often contribute differently to the model training. Therefore, we design a transformer-based keyword refiner that increases the weights of keywords more relevant to visual cues while reducing the weights of those less relevant or unrelated to visual cues. Specifically, we extract keyword sequence from a pseudo-labeled sentence, \ie, $\hat{\mathcal{Y}}^k = \{\hat{\bm{y}}_{t'}^k \in \mathbb{R}^{N_{voc}} | 1 \le t' \le N_{word}\}$, where $\hat{\bm{y}}_{t'}^k$ is the one-hot vector of the $t'$-th keyword. We reshape these vectors to a temporary matrix $\hat{\mathbf{F}}^k \in \mathbb{R}^{N_{word} \times N_{voc}}$ and apply a fully-connected layer to reduce its dimension, leading to keyword feature matrix $\mathbf{F}^k \in \mathbb{R}^{N_{word} \times d}$. Essentially, the keyword refiner is composed of $L'$ transformer blocks, and each block includes two multi-head attention layers, three layer normalizations, and one feed-forward network. The output $\bm{F}_{l'-1}^k\in \mathbb{R}^{N_{word} \times d}$ of the previous ($l'$-1)-th block is the input of the current $l'$-th block, and the input of the first block is video feature $\mathbf{V}_L$ and keyword feature $\mathbf{F}^k$. Formally, the output $\mathbf{F}_{l'}^k$ of the $l'$-th block is obtained by
\begin{equation}
	\begin{aligned}
		\label{eq:keyword_fea_output}
		\hat{\mathbf{F}}_{l'}^k & = \mathrm{LayerNorm}(\mathbf{F}_{l'-1}^k + \mathrm{MultiHead}(\mathbf{F}_{l'-1}^k, \mathbf{F}_{l'-1}^k, \mathbf{F}_{l'-1}^k)), \\
		\tilde{\mathbf{F}}_{l'}^k & = \mathrm{LayerNorm}(\hat{\mathbf{F}}_{l'}^k + \mathrm{MultiHead}(\hat{\mathbf{F}}_{l'}^k, \bm{V}_L, \bm{V}_L)), \\
		\mathbf{F}_{l'}^k & = \mathrm{LayerNorm}(\tilde{\mathbf{F}}_{l'}^k + \mathrm{FFN}(\tilde{\mathbf{F}}_{l'}^k)),
	\end{aligned}
\end{equation}
where the temporary feature $\hat{\mathbf{F}}_{l'}^k$ is obtained by adopting the self-attention for $\mathbf{F}_{l'-1}^k$, and $\tilde{\mathbf{F}}_{l'}^k$ is obtained by adopting the cross attention for keyword feature $\hat{\mathbf{F}}_{l'}^k$ and video feature $\bm{V}_L$ to align the video-keyword semantics. Therefore, the final output of the keyword refiner is pseudo-labeled keyword feature $\mathbf{F}_{L'}^k \in \mathbb{R}^{N_{word} \times d}$. By leveraging refined keyword features, the model is empowered to capture the global textual context that aligns with corresponding visual cues.

\textbf{Sentence Decoder}. To generate sentences of video, we first embed both video feature $\bm{V}_L$ and pseudo-labeled keyword feature $\mathbf{F}_{L'}^k$ into sentence semantic space by a fully-connected layer, which results in an embedded video feature matrix $\bm{V'} \in \mathbb{R}^{len_s \times d}$ and an embedded keyword feature matrix $\mathbf{F'}^k\in \mathbb{R}^{len_s \times d}$. They are fed into  $L''$ transformer blocks, and each block consists of two multi-head attention layers, one layer normalizations, and one feed-forward network. The output $\bm{H}_{l''-1} \in \mathbb{R}^{len_s \times d}$ of the previous ($l''$-1)-th block is the input of the current $l''$-th block, and the input of the first block is video feature $\mathbf{V'}$ and keyword feature $\mathbf{F'}^k$. Formally, the output $\mathbf{H}_{l''}$ of the $l''$-th block is obtained by
\begin{equation}
	\begin{aligned}
		\label{eq:decode_fea_output}
		\bm{H}_{l''}^v   & = \mathrm{MultiHead}(\bm{H}_{l''-1}, \bm{V'}, \bm{V'}), \\
		\bm{H}_{l''}^k   & = \mathrm{MultiHead}(\bm{H}_{l''-1}, \bm{F'}^k, \bm{F'}^k), \\
		\mathbf{G}_{l''} & = \sigma(\bm{W}_g\cdot[\bm{H}_{l''}^v; \bm{H}_{l''}^k]), \\
		\hat{\bm{H}}_{l''}  & = \mathbf{G}_{l''} \circ \bm{H}_{l''}^v + (\bm{I}-\mathbf{G}_{l''}) \circ \bm{H}_{l''}^k, \\
		\bm{H}_{l''}     & = \mathrm{LayerNorm}(\hat{\bm{H}}_{l''} + \mathrm{FFN}(\hat{\bm{H}}_{l''})),
	\end{aligned}
\end{equation}
where $\circ$ is element-wise product, $[\cdot; \cdot]$ denotes the concatenation along feature dimension, $\sigma(\cdot)$ is the sigmoid function, $\bm{I}\in \mathbb{R}^{len_s \times d}$ is an all-one matrix, and $\bm{W}_g\in \mathbb{R}^{len_s\times 2len_s}$ is a learnable weight matrix. Here, $\mathbf{G}_{l''}\in \mathbb{R}^{len_s \times d}$ is gated matrix which considers both video context feature $\bm{H}_{l''}^v\in \mathbb{R}^{len_s \times d}$ and text context feature $\bm{H}_{l''}^v\in \mathbb{R}^{len_s \times d}$, while $\hat{\mathbf{H}}_{l''}\in \mathbb{R}^{len_s \times d}$ is gated fusion matrix that considers the different contributions of visual cues and text semantics. Therefore, the final output of the video-text context feature $\bm{H}_{L''} \in \mathbb{R}^{len_s \times d}$. 

In the end, we feed the feature $\bm{H}_{L''}$ to a fully-connected layer with a softmax function $\mathrm{softmax}(\cdot)$, resulting in a word probability matrix $\bm{P}\in \mathbb{R}^{len_s\times N_{voc}}$, \ie,
\begin{equation}
	\label{eq:word_prob}
	\mathbf{P} = \mathrm{softmax}(\mathbf{H}_{L''} \cdot \mathbf{W}_p ),
\end{equation}
where $\bm{W}_p\in \mathbb{R}^{d \times N_{voc}}$ is a learnable weight matrix. With the keyword refiner, it enables considering the global context of both video features and text features, which benefits the prediction of possible words in video captions.

Now we can generate the sentence $\mathcal{Y}^\prime = \{\mathbf{y}_t^\prime \in \mathbb{R}^{N_{voc}}| 1 \le t \le len_s\}$ according to the word probability matrix $\mathbf{P}$. First, the entry with the maximum value in each row of $\mathbf{P}$ is set to 1, and other entries are set to 0. Second, we sequentially look up the words in the position where the entry is 1 in vocabulary to constitute the sentence. 

\subsection{Loss Function}
There are two losses to be optimized by our captioning model, including sentence loss $\mathcal{L}_{sen}$ and word loss $\mathcal{L}_{word}$. 

\textbf{Sentence Loss}. We adopt the cross-entropy to compute the sentence loss, \ie,
\begin{equation}
	\label{eq:loss_sen}
	\mathcal{L}_{sen} = -\frac{1}{len_s}\sum_{t=1}^{len_s} (\hat{\mathbf{y}}_t^\top + \mathbf{y}_t^\top )\log(\mathbf{p}_t),
\end{equation}
where $\{\mathbf{y}_t,\hat{\mathbf{y}}_t\}\in \mathbb{R}^{N_{voc}}$ are the one-hot vectors of the $t$-th word in ground-truth and pseudo-labeled sentence, respectively; $\mathbf{p}_t\in \mathbb{R}^{N_{voc}}$ is the $t$-th word probability vector in $\mathbf{P}$. 

\textbf{Word Loss}. First, we embed the pseudo-labeled keyword feature matrix $\mathbf{F}_{L'}^k\in \mathbb{R}^{N_{word}\times d}$ into a feature vector $\bm{f}^k \in \mathbb{R}^d$ by a fully-connected layer with the max pooling operation. Then, to keep the semantic consistency between pseudo-labeled keyword and ground-truth keyword, we adopt the cosine similarity to compute the word loss between the pseudo-labeled keyword feature vector $\hat{\bm{f}}^k \in \mathbb{R}^d$ and the ground-truth keyword feature vector $\bm{f}^k \in \mathbb{R}^d$, \ie,
\begin{equation}
	\label{eq:loss_word}
	\mathcal{L}_{word} =  1 - \frac{\hat{\bm{f}}^k\cdot \bm{f}^k}{\|\hat{\bm{f}}^k\|_2\cdot \|\bm{f}^k\|_2},
\end{equation}
where $\|\cdot\|_2$ denotes $\ell_2$-norm, and $\bm{f}^k$ is obtained by applying a pretrained language model SBERT \cite{reimers-emnlp2019-sbert} to the keyword sequence of ground-truth sentence.

Therefore, the total loss is to sum up the sentence loss and the word loss, \ie, $\mathcal{L} = \mathcal{L}_{sen} + \mathcal{L}_{word}$.

% ------------------------ Experiments -----------------------------
\section{Experiments}
\label{test}
All experiments were performed on a server equipped with four NVIDIA TITAN Xp graphics cards. The codes are compiled for PyTorch 1.8, Python 3.8, and CUDA 10.2.

\subsection{Datasets}
We conduct experiments on three benchmarks, including MSVD\cite{chen-acl2011-msvd}, MSR-VTT\cite{xu-cvpr2016-msr-vtt}, and VATEX\cite{wang-iccv2019-vatex}. Their statistics are summarized in Table~\ref{tbl:dataset}. Note that we examine the model performance on the test set.

% ----------------- Dataset Statistics -----------
\begin{table}[!t]
	\centering
	\caption{Statistics of Dataset.}
	\label{tbl:dataset}
	\begin{tabular}{ l  r  r  r  r }
		\toprule[0.75pt]
		Dataset & Video & Training & Validation  & Test\\
		\midrule[0.5pt]
		MSVD\cite{chen-acl2011-msvd}    & 1,970  & 1,200  & 100  & 670  \\
		MSR-VTT\cite{xu-cvpr2016-msr-vtt} & 10,000 & 6,513  & 497  & 2,990 \\
		VATEX\cite{wang-iccv2019-vatex}   & 34,991 & 25,991 & 3,000 & 6,000 \\
		\toprule[0.75pt]
	\end{tabular}
	%}
\end{table}

\textbf{MSVD} (Microsoft Research Video Description)\cite{chen-acl2011-msvd} \footnote{\url{https://www.cs.utexas.edu/users/ml/clamp/videoDescription}}. This corpus has 122K descriptions for 2,089 video clips originally, but leaves 1,970 clips with 85,550 English captions due to the volatility of YouTube website. Each clip has a duration of 10 to 25 seconds, and has roughly 40 descriptions. %There are 1200 clips for training, 100 for validation, and 670 for test.

\textbf{MSR-VTT} (Microsoft Research Video to Text)\cite{xu-cvpr2016-msr-vtt}\footnote{\url{https://www.kaggle.com/datasets/vishnutheepb/msrvtt}}. It provides 10,000 video clips with 41.2 hours and 200K clip-sentence pairs from 20 categories. Each clip has about 20 English sentences annotated by 1,327 Amazon Mechanical Turk (AMT) workers. It is collected by 257 popular video queries, and has about 29,000 unique words in all captions. %The splits are 6,513 clips for training, 497 clips for validation and 2,990 clips for test.

\textbf{VATEX}\cite{wang-iccv2019-vatex}\footnote{\url{https://eric-xw.github.io/vatex-website}}. It is large-scale multilingual video description dataset, which contains over 41,250 videos (6278 for private test) from YouTube website and 825,000 lexically-rich captions in both English and Chinese. The clip-sentence pairs are annotated with multiple unique sentences (about 10). It contains more comprehensive yet representative video content, covering 600 fine-grained human activities in total. Due to some unavailable links, we use 22,765 videos for training, 2,644 videos for validation, and 5,244 videos for test. 

\subsection{Evaluation Metrics}
We adopt the commonly used four metrics \cite{ye-cvpr2022-hmn} to evaluate the caption quality, including BLEU, METEOR, ROUGE-L, and CIDEr-D. Our records are made by using the standard evaluation code on MSCOCO server. These metrics serve as tools for evaluating machine translations against human ones, and they exhibit varying degrees of correlation with human assessment. 
% BLEU \cite{papineni-acl2002-bleu}, METEOR \cite{denkowski-smt2014-meteor}, ROUGE-L \cite{lin-acl2004-rouge}, and CIDEr-D \cite{vedantam-cvpr2015-cider}. 
% MSCOCO server \cite{chen-corr2015-coco}

Among them, BLEU records the co-occurrences of n-grams between the candidate and reference sentences; METEOR makes an alignment between the words in candidate and reference sentences, and is computed by minimizing the chunk number of contiguous and identically ordered tokens in the sentence pair; ROUGE-L is a measure based on the Longest Common Subsequence (LCS) that includes the words shared by two sentences occurring in the same order; CIDEr-D computes the cosine similarity of candidate and reference sentences, by performing a Term Frequency Inverse Document Frequency (TF-IDF) weighting for each n-gram. %Among them, CIDEr is considered more consistent with human evaluation \cite{zheng-cvpr2020-SAAT}.

\subsection{Experimental Setup}
\label{sec:exp_set}
\textbf{Training}. In pseudo-labeling module, we pretrain the language model XLNet on the MSCOCO image database with 500,000 sentences for training, 10,000 sentences for validation, and 50,000 sentences for test. We choose the optimal language model when the validation loss is lowest. Meanwhile, we use the created synthetic data to finetune XLNet, yielding the token-level classifier that decides which token should be edited in sentence. The number of keywords $N_{word}$ is set to 4 for MSVD\cite{chen-acl2011-msvd}, 5 for MSR-VTT\cite{xu-cvpr2016-msr-vtt}, and 7 for VATEX\cite{wang-iccv2019-vatex}. 

For video encoder, following \cite{ye-cvpr2022-hmn}, we use the Inception ResNetv2 \cite{szegedy-aaai2017-inception-v4} model pretrained on ImageNet \cite{russakovsky-ijcv2015-imagenet} database to extract appearance features (the output of ``avg pool" layer) and the dimension is $d_a=1536$; we use the C3D \cite{hara-cvpr2018-c3d} model pretrained on Sports1M \cite{karpathy-cvpr2014-sports} database to extract motion features (the output of ``fc6" layer, 16-frame clip with 8 overlapped frames) and the dimension is $d_m=2048$; we use the Faster-RCNN \cite{ren-tpami2017-faster-rcnn} model pretrained on Visual Genome \cite{krishna-ijcv2017-visual-genome} database to extract object features (the output of ``fc7" layer, 10 objects in the 8-th frame of a 16-frame clip without overlapping) and the dimension is $d_o = 2048$. As the number of these features varies with different videos, we set the maximum number of features to a fixed number, \ie, $\{N, N_{obj}\}$ are set to $\{20,20\}$ for MSVD\cite{chen-acl2011-msvd}, $\{30, 40\}$ for MSR-VTT\cite{xu-cvpr2016-msr-vtt}, and $\{30,30\}$ for VATEX\cite{wang-iccv2019-vatex}. If the number of original features is larger than the maximum number, we simply adopt the average sampling strategy. Moreover, these features are mapped to a low-dimensional data space by a fully-connected layer, and the dimension is reduced to $d=768$. For captions, we remove all punctuations and convert all letters to lower-case ones. Each caption contains a start label ``BOS" and an end label ``EOS''. We truncate those captions longer than 20 words, and pad zeros for those less than 20 words. The vocabulary is composed of the words whose frequency is larger than 1, and the vocabulary size $N_{voc}$ is 12,800 for MSVD\cite{chen-acl2011-msvd}, 28,485 for MSR-VTT\cite{xu-cvpr2016-msr-vtt}, and 21,784 for VATEX\cite{wang-iccv2019-vatex}. The number of transformer blocks $L/L'/L''$ are set to $\{1/1/8, 3/2/6, 2/3/4\}$ and $\{1/2/4, 3/3/6, 2/3/6\}$ on the datasets \{MSVD, MSR-VTT, VATEX\}, in fully-supervised and few-supervised scenario, respectively.

We train the model by the Adam \cite{kingma-iclr2015-adam} optimizer with an initial learning rate $1e-4$ and weight decay 0.5. The head number is 8 in multi-head attention, batch size is 128, and the maximum epoch is 35. We adopt the early stopping \cite{zheng-cvpr2020-saat} strategy, \ie, the model converges when the CIDEr-D score does not rise in continuos five epochs. 

\textbf{Inference}. Given a test video, we extract its appearance feature, motion feature, and object feature as in training. These features are fed into the captioning module to yield the word probability matrix, which identifies the words to appear in the corresponding caption from the vocabulary.

% -------------- Results of full supervision ------
% ---------------------- Results on MSVD and MSR-VTT (supervised) ----------------
\begin{table*}[!t]
	\centering
	\caption{Performance comparisons on MSVD\cite{chen-acl2011-msvd} and MSR-VTT\cite{xu-cvpr2016-msr-vtt} with full supervision. Here, `B' is BLEU, `M' is METOR, `R' is ROUGE-L, `C' is CIDEr-D.}
	\label{table:msvd_vtt_full}
	\resizebox{\textwidth}{!}{
		\setlength{\tabcolsep}{0.8mm}{ 
			\begin{tabular}{lr ccc c cccc c cccc}
				\toprule[0.75pt]
				\multirow{2}{*}{Method} & \multirow{2}{*}{Venue} & \multicolumn{4}{c}{Feature} & \multicolumn{5}{c}{MSVD\cite{chen-acl2011-msvd}} & \multicolumn{4}{c}{MSR-VTT\cite{xu-cvpr2016-msr-vtt}}  \\ \cmidrule[0.5pt]{3-5}  \cmidrule[0.5pt]{7-10} \cmidrule[0.5pt]{12-15}
				& & Appear. & Motion & Object & & B@4$\uparrow$ & M$\uparrow$ & R$\uparrow$ & C$\uparrow$ & & B@4$\uparrow$ & M$\uparrow$ & R$\uparrow$ & C$\uparrow$  \\ 
				\midrule[0.5pt]
			%	MARN\cite{pei-cvpr2019-marn}         &CVPR'19 &RN-101 &C3D         &-           & &48.6                   &35.1                &71.9                &92.2                & &10.4                &28.1                &60.7                &47.1                \\ 
			%	OA-BTG\cite{zhang-cvpr2019-oa-btg}   &CVPR'19 &RN-200 &-           &MR   & &52.5                   &34.1                &71.3                &88.7                 & &42.0                &28.2                &61.6                &48.7                \\ 
				%POS-CG\cite{wang-iccv2019-pos-cg}    &ICCV'19 &IRv2       &OF &-           & &52.5                   &34.1                &71.3                &88.7                 & &42.0                &28.2                &61.6                &47.5                \\ 
			%	MGSA\cite{chen-aaai2019-mgsa}        &AAAI'19 &IRv2       &C3D         &-           & &53.4                   &35.0                &-                   &86.7                 & &42.4                &27.6                &-                   &47.5                \\ 
				%STG-KD\cite{pan-cvpr2020-stg-kd}     &CVPR'20 &RN-101 &I3D         &FR & &52.2                   &36.9                &73.9                &93.0                 & &40.5                &28.3                &60.9                &47.1                \\ 
				SAAT\cite{zheng-cvpr2020-saat}       &CVPR'20 &IRv2       &C3D         &FR & &46.5                   &33.5                &69.4                &81.0                 & &40.5                &28.2                &60.9                &49.1                \\ 
				ORG-TRL\cite{zhang-cvpr2020-org-trl} &CVPR'20 &IRv2       &C3D         &FR & &54.3                   &36.4                &73.9                &95.2                 & &43.6                &28.8                &62.1                &50.9                \\ 
				SGN\cite{ryu-aaai2021-sgn}           &AAAI'21 &RN-101 &C3D         &-           & &52.8                   &35.5                &72.9                &94.3                 & &43.6                &28.3                &60.8                &49.5                \\ 
				MGRMP\cite{chen-iccv2021-mgrmp}      &ICCV'21 &IRv2       &C3D         &            & &55.8                   &36.9                &74.5                &98.5                 & &41.7                &28.9                &62.1                &51.4                \\ 
				HRNAT\cite{gao-tip2022-hrnat}        &TIP'22  &IRv2       &I3D         &-           & &55.7                   &36.8                &74.1                &98.1                 & &42.1                &28.0                &61.6                &48.2                \\ 
				LSRT\cite{li-tip2022-lstg}           &TIP'22  &IRv2       &C3D         &FR & &55.6                   &37.1                &73.5                &98.5                 & &42.6                &28.3                &61.0                &49.5                \\ 
				HMN\cite{ye-cvpr2022-hmn}            &CVPR'22 &IRv2       &C3D         &FR & &59.2                   &37.7                &75.1                &104.0                & &43.5                &29.0                &\underline{62.7}    &51.5                \\ 
				MAN\cite{jing-tmm2023-man}           &TMM'23  &RN-101   &RNX-101 &-           & &59.7                   &37.3                &74.3                &101.5                & &41.3                &28.0                &61.4                &49.8                \\ 
				TextKG\cite{gu-cvpr2023-textkg}      &CVPR'23 &IRv2       &C3D         &FR & &\textbf{60.8}          &\underline{38.5}    &\underline{75.1}    &\underline{105.2}    & &\underline{43.7}    &\underline{29.6}    &62.4                &\underline{52.4}    \\ 
				\toprule[0.5pt]
				Ours                                 &        &IRv2       &C3D         &FR & &\underline{60.1} &\textbf{39.3} &\textbf{76.2} &\textbf{107.2} & &\textbf{44.8} &\textbf{30.5} &\textbf{63.4} &\textbf{53.5} \\ 
				&        &        &          &  & & -0.7 & +0.8 &+1.1 &+2.0 & &+1.1 &+0.9 &+0.7 &+1.1 \\ 
				\toprule[0.75pt]
			\end{tabular}
		}
	}
\end{table*}

% ---------------------- Results on VATEX (supervised) ----------------
\begin{table}[!t]
	\centering
	\caption{Performance comparisons on VATEX\cite{wang-iccv2019-vatex} with full supervision.}
	\label{table:vatex_full}
	\setlength{\tabcolsep}{0.6mm}{ 
		\begin{tabular}{lr ccc c cccc}
			\toprule[0.75pt]
			\multirow{2}{*}{Method} & \multirow{2}{*}{Venue} & \multicolumn{4}{c}{Feature} & \multicolumn{4}{c}{VATEX\cite{wang-iccv2019-vatex}} \\ 
			\cmidrule[0.5pt]{3-5}  \cmidrule[0.5pt]{7-10}
			& & Appear. & Motion & Object & & B@4$\uparrow$ & M$\uparrow$ & R$\uparrow$ & C$\uparrow$  \\ 
			\midrule[0.5pt]
			ORG-TRL\cite{zhang-cvpr2020-org-trl} &CVPR’20 &IRv2   &C3D     &FR & &32.1                &22.2                &48.9                &49.7       \\             
			HRNAT\cite{gao-tip2022-hrnat}        &TIP’22  &IRv2   &I3D     &-           & &32.5                &22.3                &49.0                &\underline{50.7}       \\             
			MAN\cite{jing-tmm2023-man}           &TMM’23  &RN   &RNX &-           & &\underline{32.7}    &\underline{22.4}    &\underline{49.1}    &48.9   \\ 
			\toprule[0.5pt]                      
			Ours       &         &IRv2   &C3D    &FR & &\textbf{34.3} &\textbf{23.5} &\textbf{49.5} &\textbf{53.3}\\   
			&         &   &    &  & & +1.6 & +1.1 &+0.4 &+2.6 \\    
			\toprule[0.75pt]      
		\end{tabular}
	}
\end{table}

% ------------------
\subsection{Compared Methods}
\label{sec:exp_compare}
To investigate the performance of the proposed PKG approach, we compare a series of competing methods. Most of them belong to encoder-decoder architecture, which often adopts two-dimensional CNN, such as ResNet (RN) \cite{he-cvpr2016-resnet} and Inception ResNetv2 (IRv2) \cite{szegedy-aaai2017-inception-v4}, and three-dimensional CNN, such as I3D \cite{carreira-cvpr2017-I3D}, C3D \cite{hara-cvpr2018-c3d}, and ResNeXt (RNX), to capture appearance feature and motion features respectively. Some of them use object detector such as Faster-RCNN (FR) \cite{ren-tpami2017-faster-rcnn} to extract object feature. Their decoders often adopt LSTM or GRU \cite{cho-emnlp2014-gru}. They can be roughly divided into two groups as follows. 
% and Optical Flow (OF, \eg, TVL1-flow) \cite{perez-ipol2013-TV-L1}
%  ResNeXt (RNX) \cite{xie-arxiv2016-resnext}   %Mask-RCNN (MR) \cite{he-iccv2017-mask-rcnn} 

In the \emph{attention-based} group, the typical methods contain SAAT (Syntax-Aware Action Targeting) \cite{zheng-cvpr2020-saat}, SGN (Semantic Grouping Network) \cite{ryu-aaai2021-sgn}, MGRMP (Motion Guided Region Message Passing) \cite{chen-iccv2021-mgrmp}, HRNAT (Hierarchical Representation Network with Auxiliary Tasks) \cite{gao-tip2022-hrnat}, and MAN (Memory-based Augmentation Network) \cite{jing-tmm2023-man}. In the \emph{graph-based} group, the typical methods contain ORG-TRL (Object Relational Graph with Teacher-Recommended Learning) \cite{zhang-cvpr2020-org-trl}, and LSRT (Long Short-Term Graph) \cite{li-tip2022-lstg}.
% attention: MARN (Motion Guided Spatial Attention) \cite{pei-cvpr2019-marn},  MGSA (Motion Guided Spatial Attention) \cite{chen-aaai2019-mgsa}, POS-CG (Part-of-Speech with Cross-Gating block) \cite{wang-iccv2019-pos-cg},
% graph:  OA-BTG (Object-aware Aggregation with Bidirectional Temporal Graph)\cite{zhang-cvpr2019-oa-btg},STG-KD (Spatio-Temporal Graph with Knowledge Distillation) \cite{pan-cvpr2020-stg-kd}, 

In addition, we compare two transformer-based SOTA alternatives, including HMN (Hierarchical Modular Network) \cite{ye-cvpr2022-hmn} and TextKG (Text with Knowledge Graph) \cite{gu-cvpr2023-textkg}.

% ----------------- Few Supervision ------------
% ---------------------- Results on three benchmarks (few-supervised) ----------------
\begin{table*}[!t]
	\centering
	\caption{Performance comparisons on three benchmarks with few supervision (1 GT sentence).}
	\label{table:1sen_few}
	\resizebox{\textwidth}{!}{
	\setlength{\tabcolsep}{0.6mm}{ 
		\begin{tabular}{lr ccc ccc c cccc c cccc c cccc}
			\toprule[0.75pt]
			\multirow{2}{*}{Method} & \multirow{2}{*}{Venue} & \multicolumn{4}{c}{Feature} & \multicolumn{2}{c}{Sentence} & \multirow{2}{*}{}  & \multicolumn{5}{c}{MSVD\cite{chen-acl2011-msvd}} & \multicolumn{5}{c}{MSR-VTT\cite{xu-cvpr2016-msr-vtt}} & \multicolumn{4}{c}{VATEX\cite{wang-iccv2019-vatex}}  \\ 
			\cmidrule[0.5pt]{3-5} \cmidrule[0.5pt]{7-8}  \cmidrule[0.5pt]{10-13} \cmidrule[0.5pt]{15-18} \cmidrule[0.5pt]{20-23}
			& & Appear. & Motion & Object & & GT & Pse &  & B@4$\uparrow$ & M$\uparrow$ & R$\uparrow$ & C$\uparrow$ & & B@4$\uparrow$ & M$\uparrow$ & R$\uparrow$ & C$\uparrow$ & & B@4$\uparrow$ & M$\uparrow$ & R$\uparrow$ & C$\uparrow$  \\ 
			\midrule[0.5pt]
			SAAT\cite{zheng-cvpr2020-saat} &CVPR'20  &IRv2   &C3D &FR & &1 &0  &&10.6 &20.4 &57.1 &10.5 & 
			&13.9  &13.7  &46.3  &15.9 & &20.5 &11.2 &31.5 &22.8  \\ 
			SGN\cite{ryu-aaai2021-sgn}     &AAAI'21  &RN-101 &C3D &-  & &1 &0  &&11.8 &20.8 &58.0 &13.0 & 
			&20.3  &15.1  &47.6  &20.7 & &22.4 &12.7 &34.4 &24.5  \\ 
			HRNAT\cite{gao-tip2022-hrnat}  &TIP'22   &IRv2   &I3D &-  & &1 &0  &&16.5 &22.3 &60.4 &17.9 & 
			&20.4  &16.3  &48.3  &18.8 & &24.0 &14.9 &40.9 &31.4  \\ 
			HMN\cite{ye-cvpr2022-hmn}      &CVPR'22  &IRv2   &C3D &FR & &1 &0  &&18.9 &24.2 &60.7 &19.3 & 
			&22.3  &17.3  &49.2  &23.2 & &24.6 &15.7 &41.5 &32.1  \\ 
			MAN\cite{jing-tmm2023-man}     &TMM'23&RN-101&RNX-101 &-  & &1 &0  &&\underline{20.5} &\underline{25.3} &\underline{61.4} &\underline{20.9} & &\underline{24.2} &\underline{18.2} &\underline{49.6} &\underline{24.8} & 
			&\underline{25.1}   &\underline{16.1} &\underline{42.6} &\underline{34.2} \\
			\toprule[0.5pt]
			Ours     &        &IRv2   &C3D   &FR & &1 &0 &&\textbf{23.3} &\textbf{27.1}  &\textbf{63.5} &\textbf{21.7} & 
			&\textbf{26.2}    &\textbf{18.8}  &\textbf{50.3}   &\textbf{26.4} &
			&\textbf{26.9}    &\textbf{17.1}  &\textbf{43.6}   &\textbf{36.8} \\ 
			& & & & & & & & &+2.8 &+1.8 &+2.1 &+0.8 & &+2.0 &+0.6 &+0.7 &+1.6 & &+1.8 &+1.0 &+1.0 &+2.6  \\ 
			\toprule[0.5pt]
			SAAT\cite{zheng-cvpr2020-saat} &CVPR'20  &IRv2   &C3D &FR & &2 &0  &&26.5 &26.2 &59.2 &29.9 & 
			&17.7 &17.8 &47.3 &13.3 & &21.2 &12.5 &37.8 &31.3 \\ 
			SGN\cite{ryu-aaai2021-sgn}     &AAAI'21  &RN-101 &C3D &-  & &2 &0  &&25.9 &25.0 &62.1 &34.7 & 
			&23.5 &18.4 &49.2 &25.4 & &22.7 &13.4 &39.2 &34.1 \\ 
			HRNAT\cite{gao-tip2022-hrnat}  &TIP'22   &IRv2   &I3D &-  & &2 &0  &&28.7 &27.6 &62.5 &47.8 & 
			&24.6 &19.1 &50.3 &24.6 & &25.8 &15.8 &42.7 &36.2 \\ 
			HMN\cite{ye-cvpr2022-hmn}      &CVPR'22  &IRv2   &C3D &FR & &2 &0  &&29.1 &29.3 &63.2 &\underline{51.3} & 
			&26.4 &19.2 &50.8 &\underline{27.5} & &26.3 &16.5 &42.8 &\underline{38.2} \\ 
			MAN\cite{jing-tmm2023-man}     &TMM'23&RN-101&RNX-101 &-  & &2 &0  &&\underline{29.9} &\underline{30.2} &\underline{63.5} &50.3 & 
			&\underline{26.6} &\underline{19.8} &\underline{51.2} &26.8 &
			&\underline{26.6} &\underline{17.4} &\underline{43.2} &37.1 \\
			\toprule[0.5pt]
			Ours     &        &IRv2   &C3D   &FR & &1 &1 &&\textbf{31.3} &\textbf{31.0} &\textbf{64.4} &\textbf{52.8} & 
			&\textbf{27.6} &\textbf{20.2} &\textbf{51.9} &\textbf{28.7} &
			&\textbf{27.6} &\textbf{18.4} &\textbf{44.1} &\textbf{40.4} \\ 
			& & & & & & & & &+1.4 &+0.8 &+0.9 &+1.5 & &+1.0 &+0.4 &+0.7 &+1.2 & &+1.0 &+1.0 &+0.9 &+2.2 \\ 
			\toprule[0.75pt]
		\end{tabular}
	}
}
\end{table*}

\begin{table*}[!t]
	\centering
	\caption{Performance comparisons on three benchmarks with few supervision (3 GT sentences).}
	\label{table:3sen_few}
	\resizebox{\textwidth}{!}{
	\setlength{\tabcolsep}{0.5mm}{ 
		\begin{tabular}{lr ccc ccc c cccc c cccc c cccc}
			\toprule[0.75pt]
			\multirow{2}{*}{Method} & \multirow{2}{*}{Venue} & \multicolumn{4}{c}{Feature} & \multicolumn{2}{c}{Sentence} & \multirow{2}{*}{}  & \multicolumn{5}{c}{MSVD\cite{chen-acl2011-msvd}} & \multicolumn{5}{c}{MSR-VTT\cite{xu-cvpr2016-msr-vtt}} & \multicolumn{4}{c}{VATEX\cite{wang-iccv2019-vatex}}  \\ 
			\cmidrule[0.5pt]{3-5} \cmidrule[0.5pt]{7-8}  \cmidrule[0.5pt]{10-13} \cmidrule[0.5pt]{15-18} \cmidrule[0.5pt]{20-23}
			& & Appear. & Motion & Object & & GT & Pse &  & B@4$\uparrow$ & M$\uparrow$ & R$\uparrow$ & C$\uparrow$ & & B@4$\uparrow$ & M$\uparrow$ & R$\uparrow$ & C$\uparrow$ & & B@4$\uparrow$ & M$\uparrow$ & R$\uparrow$ & C$\uparrow$  \\ 
			\midrule[0.5pt]
			SAAT\cite{zheng-cvpr2020-saat} &CVPR'20  &IRv2   &C3D &FR & &3 &0  &&30.9 &26.3 &61.9 &54.0 & 
			&24.8 &19.5 &48.5 &24.1 & &23.5 &13.8 &39.1 &33.5  \\ 
			SGN\cite{ryu-aaai2021-sgn}     &AAAI'21  &RN-101 &C3D &-  & &3 &0  &&31.8 &29.4 &63.2 &56.2 & 
			&27.4 &20.3 &50.4 &30.4 & &25.1 &14.7 &42.6 &36.2  \\ 
			HRNAT\cite{gao-tip2022-hrnat}  &TIP'22   &IRv2   &I3D &-  & &3 &0  &&31.6 &30.3 &63.8 &57.5 & 
			&28.9 &21.6 &51.3 &31.1 & &28.7 &16.9 &43.3 &39.4  \\ 
			HMN\cite{ye-cvpr2022-hmn}      &CVPR'22  &IRv2   &C3D &FR & &3 &0  &&31.8 &30.9 &65.1 &\underline{61.7} & 
			&30.3 &22.3 &52.4 &\underline{32.9} & &29.4 &17.2 &43.7 &\underline{41.6}  \\ 
			MAN\cite{jing-tmm2023-man}     &TMM'23&RN-101&RNX-101 &-  & &3 &0  &&\underline{32.4} &\underline{31.7} &\underline{65.8} &60.3  &
			&\underline{31.2} &\underline{23.8} &\underline{53.1} &32.5 & 
			&\underline{30.2} &\underline{18.3} &\underline{44.7} &40.7 \\
			\toprule[0.5pt]
			Ours     &        &IRv2   &C3D   &FR & &3 &0 &&\textbf{34.6} &\textbf{32.5} &\textbf{67.4} &\textbf{65.1} & 
			&\textbf{34.6} &\textbf{24.3} &\textbf{54.9} &\textbf{37.9} &
			&\textbf{31.1} &\textbf{19.1} &\textbf{45.5} &\textbf{44.3} \\ 
			& & & & & & & & &+2.2 &+0.8 &+1.6 &+3.4 & &+3.4 &+0.5 &+1.8 &+5.0 & &+0.9 &+0.8 &+0.8 &+2.7 \\ 
			\toprule[0.5pt]
			SAAT\cite{zheng-cvpr2020-saat} &CVPR'20  &IRv2   &C3D &FR & &6 &0  &&38.2 &28.3 &63.5 &65.1 & 
			&32.1 &22.2 &52.2 &36.9 & &24.5 &14.6 &41.2 &37.3 \\ 
			SGN\cite{ryu-aaai2021-sgn}     &AAAI'21  &RN-101 &C3D &-  & &6 &0  &&38.8 &29.3 &64.1 &69.8 & 
			&33.8 &23.4 &53.8 &38.3 & &28.3 &15.7 &43.8 &40.9 \\ 
			HRNAT\cite{gao-tip2022-hrnat}  &TIP'22   &IRv2   &I3D &-  & &6 &0  &&41.8 &30.9 &64.8 &71.4 & 
			&34.9 &23.9 &54.2 &40.3 & &29.4 &17.4 &44.2 &41.1 \\ 
			HMN\cite{ye-cvpr2022-hmn}      &CVPR'22  &IRv2   &C3D &FR & &6 &0  &&42.3 &31.7 &\underline{66.8} &\underline{73.4} &
			&\textbf{36.5} &\underline{24.8} &\underline{54.9} &\underline{41.6} & 
			&\underline{31.2} &\underline{18.7} &\underline{46.1} &\underline{44.1} \\ 
			MAN\cite{jing-tmm2023-man}     &TMM'23&RN-101&RNX-101 &-  & &6 &0  &&\underline{42.9} &\underline{32.5} &65.8 &72.7 & 
			&\underline{35.8} &\textbf{25.1} &54.4 &41.0 & &30.4 &18.1 &45.8 &43.3 \\
			\toprule[0.5pt]
			Ours     &        &IRv2   &C3D   &FR & &3 &3 &&\textbf{43.8} &\textbf{32.9} &\textbf{68.0} &\textbf{74.9} & 
			&35.9 &24.6 &\textbf{55.6} &\textbf{42.3} &
			&\textbf{31.5} &\textbf{19.4} &\textbf{46.3} &\textbf{45.2} \\ 
			& & & & & & & & &+0.9 &+0.4 &+1.2 &+1.5 & &-0.6 &-0.5 &+0.7 &+0.7 & &+0.3 &+0.7 &+0.2  &+1.1 \\ 
			\toprule[0.75pt]
		\end{tabular}
	}
}
\end{table*}

% ----------------- Results with five GT sentences 
\begin{table*}[!t]
	\centering
	\caption{Performance comparisons on three benchmarks with few supervision (5 GT sentences).}
	\label{table:5sen_few}
	\resizebox{\textwidth}{!}{
	\setlength{\tabcolsep}{0.5mm}{ 
		\begin{tabular}{lr ccc ccc c cccc c cccc c cccc}
			\toprule[0.75pt]
			\multirow{2}{*}{Method} & \multirow{2}{*}{Venue} & \multicolumn{4}{c}{Feature} & \multicolumn{2}{c}{Sentence} & \multirow{2}{*}{} & \multicolumn{5}{c}{MSVD\cite{chen-acl2011-msvd}} & \multicolumn{5}{c}{MSR-VTT\cite{xu-cvpr2016-msr-vtt}} & \multicolumn{4}{c}{VATEX\cite{wang-iccv2019-vatex}}  \\ 
			\cmidrule[0.5pt]{3-5} \cmidrule[0.5pt]{7-8}  \cmidrule[0.5pt]{10-13} \cmidrule[0.5pt]{15-18} \cmidrule[0.5pt]{20-23}
			& & Appear. & Motion & Object & & GT & Pse &  & B@4$\uparrow$ & M$\uparrow$ & R$\uparrow$ & C$\uparrow$ & & B@4$\uparrow$ & M$\uparrow$ & R$\uparrow$ & C$\uparrow$ & & B@4$\uparrow$ & M$\uparrow$ & R$\uparrow$ & C$\uparrow$  \\ 
			\midrule[0.5pt]
			SAAT\cite{zheng-cvpr2020-saat} &CVPR'20  &IRv2   &C3D &FR & &5 &0  &&37.4 &27.1 &62.5 &63.2 & 
			&31.6 &21.2 &51.1 &36.1 & &23.7 &14.2 &40.7 &36.5  \\ 
			SGN\cite{ryu-aaai2021-sgn}     &AAAI'21  &RN-101 &C3D &-  & &5 &0  &&38.1 &28.8 &63.2 &67.4 & 
			&32.5 &22.6 &52.9 &37.9 & &27.9 &15.2 &43.2 &39.3  \\ 
			HRNAT\cite{gao-tip2022-hrnat}  &TIP'22   &IRv2   &I3D &-  & &5 &0  &&41.5 &29.7 &63.5 &69.5 & 
			&33.7 &23.3 &53.2 &39.2 & &29.1 &17.1 &43.7 &40.6  \\ 
			HMN\cite{ye-cvpr2022-hmn}      &CVPR'22  &IRv2   &C3D &FR & &5 &0  &&42.1 &30.5 &\underline{65.3} &\underline{72.8} & 
			&\underline{35.8} &24.1 &\underline{54.1} &\underline{40.9} & 
			&\underline{30.6} &\underline{18.5} &\underline{45.7} &\underline{43.8}  \\ 
			MAN\cite{jing-tmm2023-man}     &TMM'23&RN-101&RNX-101 &-  & &5 &0  &&\underline{42.6} &\underline{31.2} &64.5 &71.5 & 
			&34.9 &\underline{24.7} &53.7 &40.3 & &29.6 &17.7 &44.9 &43.1 \\
			\toprule[0.5pt]
			Ours     &        &IRv2   &C3D   &FR & &5 &0 &&\textbf{44.4} &\textbf{33.6} &\textbf{67.9} &\textbf{75.5} & 
			&\textbf{36.2} &\textbf{24.9} &\textbf{54.7} &\textbf{41.2} &
			&\textbf{31.4} &\textbf{19.2} &\textbf{46.2} &\textbf{46.9} \\ 
			& & & & & & & & &+1.8 &+2.4 &+2.6 &+2.7 & &+0.4 &+0.2 &+0.6 &+0.3 & &+0.8 &+0.7 &+0.5 &+3.1 \\ 
			\toprule[0.5pt]
			SAAT\cite{zheng-cvpr2020-saat} &CVPR'20  &IRv2   &C3D &FR & &10 &0  &&42.5 &32.3 &66.5 &67.6 & 
			&37.4 &23.2 &55.1 &42.1 & &27.3 &15.2 &43.2 &40.2 \\ 
			SGN\cite{ryu-aaai2021-sgn}     &AAAI'21  &RN-101 &C3D &-  & &10 &0  &&43.6 &33.0 &68.9 &72.5 & 
			&39.9 &24.2 &55.3 &42.8 & &30.3 &17.7 &44.4 &44.3 \\ 
			HRNAT\cite{gao-tip2022-hrnat}  &TIP'22   &IRv2   &I3D &-  & &10 &0  &&46.9 &33.7 &69.6 &76.3 &
			&40.9 &24.9 &56.4 &43.4 & &30.8 &18.3 &45.4 &43.5 \\ 
			HMN\cite{ye-cvpr2022-hmn}      &CVPR'22  &IRv2   &C3D &FR & &10 &0  &&47.4 &\textbf{35.3} &\textbf{71.8} &\textbf{78.3} &
			&\textbf{42.1} &\textbf{26.1} &\textbf{58.9} &\underline{44.9} &
			&\textbf{32.7} &\textbf{19.8} &\textbf{48.1} &\textbf{47.9} \\ 
			MAN\cite{jing-tmm2023-man}     &TMM'23&RN-101&RNX-101 &-  & &10 &0  &&\underline{48.1} &\underline{34.2} &\underline{70.9} &\underline{77.2} & 
			&\underline{41.7} &\underline{25.9} &\underline{57.7} &44.2 & &31.4 &19.1 &46.8 &46.2 \\
			\toprule[0.5pt]
			Ours     &        &IRv2   &C3D   &FR & &5 &5 &&\textbf{48.6} &34.2 &70.1 &77.2 & 
			&40.7 &25.3 &57.4 &\textbf{45.4} &
			&\underline{31.9} &\underline{19.6} &\underline{46.9} &\underline{47.2} \\ 
			& & & & & & & & &+0.5 &-1.1 &-1.7 &-1.1 & &-1.4 &-0.8 &-1.5 &+0.5 & &-0.8 &-0.2 &-1.2 &-0.7 \\ 
			\toprule[0.75pt]
		\end{tabular}
	}
}
\end{table*}

\subsection{Quantitative Results}
To explore the performance of the proposed PKG method, we report the comparison results of both fully-supervised (without pseudo-labeling module) and few-supervised scenarios. The best records are highlighted in boldface, and the second best records are underlined. Note that ``B@4", ``M", ``R", and ``C" denote BLEU@4, METEOR, ROUGE-L, and CIDEr-D score, respectively.

\textbf{Full Supervision}. The results on MSVD\cite{chen-acl2011-msvd} and MSR-VTT\cite{xu-cvpr2016-msr-vtt} with full supervision in Table~\ref{table:msvd_vtt_full} as well as that on VATEX\cite{wang-iccv2019-vatex} in Table~\ref{table:vatex_full}, respectively. 

From Table~\ref{table:msvd_vtt_full} and Table~\ref{table:vatex_full}, it can be observed that our method consistently achieves more promising performance than other alternatives across four metrics on the three benchmarks. For example, the performance is improved by 2.0\% and 2.6\% absolute CIDEr score compared to the best candidate TextKG\cite{gu-cvpr2023-textkg} and MAN\cite{jing-tmm2023-man} on MSVD\cite{chen-acl2011-msvd} and MSR-VTT\cite{xu-cvpr2016-msr-vtt}, respectively. We attribute this to the fact that the keyword refiner in captioning module is able to emphasize on those keywords more relevant to video content, and simultaneously it reduces the weights of less relevant or irrelevant keywords. Note that TextKG\cite{gu-cvpr2023-textkg} on MSVD\cite{chen-acl2011-msvd}in Table~\ref{table:msvd_vtt_full} performs well because it utilizes the external stream to model interactions between the additional knowledge and the built-in information, \eg, the salient object regions and speech transcripts. Moreover, our method outperforms MAN\cite{jing-tmm2023-man} on VATEX\cite{wang-iccv2019-vatex} in Table~\ref{table:vatex_full} is for the reason that we use transformer architecture to better model the global context while MAN\cite{jing-tmm2023-man} employs LSTM that fails to reveal the long-range temporal relations across all frames.

\textbf{Few Supervision}. The results with few supervision, \ie, 1-5 Ground-Truth (GT) sentences and the same number of Pseudo-labeled (Pse) sentences, on three benchmarks are shown in Table~\ref{table:1sen_few} (1+1), Table~\ref{table:3sen_few} (3+3), and Table~\ref{table:5sen_few} (5+5), where the two digits in parenthesis denote the numbers of ``GT" sentences and ``Pse" ones. For comparison, the top group in these tables show the results without pseudo-labeled sentence.

From these tables, we have several observations below. 
\begin{itemize}
	\item Our method with weak supervision performs better than or comparable to fully-supervised alternatives across benchmarks (bottom group), when increasing the number of available GT sentences from 1 to 5. For example, our method obtains a gain of 1.4\% in terms of BLEU@4 over the most competing MAN\cite{jing-tmm2023-man} method on MSVD\cite{chen-acl2011-msvd}, and 2.2\% absolute CIDEr-D score over the best candidate HMN\cite{ye-cvpr2022-hmn} on VATEX\cite{wang-iccv2019-vatex}, respectively, as shown in the bottom row of Table~\ref{table:1sen_few}. Meanwhile, our method behaves similarly with three GT sentences in Table~\ref{table:3sen_few}. This demonstrates our pseudo-labeling module generates higher-quality sentences that cater to video content, which helps to learn a satisfying video captioning model. 
	
	\item When using only a few GT sentences even without pseudo-labeled ones, our method enjoys more satisfying performances than the rest on all three databases (top group). For example, it has an improvement of 2.0\% in terms of BLEU@4 on MSR-VTT\cite{xu-cvpr2016-msr-vtt} and 2.6\% absolute CIDEr-D score on VATEX\cite{wang-iccv2019-vatex} over the strong baseline MAN\cite{jing-tmm2023-man}, as shown in the top group of Table~\ref{table:1sen_few}. Meanwhile, it upgrades the performance by 5.0\% and 3.1\% absolute CIDEr-D scores over the second best HMN\cite{ye-cvpr2022-hmn} on MSR-VTT\cite{xu-cvpr2016-msr-vtt} in the top group of Table~\ref{table:3sen_few} and on VATEX\cite{wang-iccv2019-vatex} in the top group of Table~\ref{table:5sen_few}, respectively. This validates the fact that the gated fusion mechanism in sentence decoder, which models the relations of visual cues and text semantics, indeed has positive impacts on the quality of generated captions.
	
	\item From Table~\ref{table:1sen_few} to Table~\ref{table:5sen_few}, the performance is progressively and greatly boosted by using more prior knowledge from one to three then five GT or additional pseudo-labeled sentences. For example, the BLUE@4 score rises from 23.3/31.3 (without/with pseudo-labeled sentences) to 34.6/43.8 then 44.4/48.6 and simultaneously CIDEr-D score rises from 21.7/52.8 to 65.1/74.9 then 75.5/77.2 on MSVD\cite{chen-acl2011-msvd}, while the CIDEr-D score upgrades from 36.8/40.4 to 44.3/45.2 then 46.9/47.2 on VATEX\cite{wang-iccv2019-vatex}. This indicates that more supervision knowledge at the cost of more human labeling cost promotes the model performance. Furthermore, the advancements in top group significantly surpasses that in bottom group, and this gap diminishes when more sentences are accessible. This implies that the leading labeled or pseudo-labeled sentences are of vital importance to the captioning quality, and the performance gains may plateau after reaching a certain threshold of sentences, \ie, beyond this saturation point, adding more sentences may not yield substantial improvements. 
	
	\item As seen from Table~\ref{table:5sen_few}, our method outperforms the best-performing candidate MAN\cite{jing-tmm2023-man} when using only five GT sentences. However, with few supervision, our approach achieves comparable or marginally inferior performance compared to the most-competing alternative HMN\cite{ye-cvpr2022-hmn}. We attribute this trend to the observation that as the number of sentences increases, the quality of pseudo-labeled sentences tends to decline in comparison to human-labeled ones due to label noise. 
	
\end{itemize}

%% ---------------------- Model size and inference speed ----------------
%\begin{table}[!t]
%	\centering
%	\caption{Model size and inference speed.}
%	\label{table:size_speed}
%	\setlength{\tabcolsep}{0.5mm}{ 
%	\begin{tabular}{l c  c c c c c c c c c c}
%			\toprule[0.75pt]
%			\multirow{2}{*}{Network} & \multirow{2}{*}{Backbone}  &\multicolumn{1}{c}{\scriptsize{$\#$Params}} & \multicolumn{3}{c}{VOCAug\cite{everingham-ijcv2015-pascal}} && \multicolumn{3}{c}{Cityscapes\cite{cordts-cvpr2016-cityscapes} }\\
%			\cmidrule{4-6} \cmidrule{8-10}			
%			&&(M) $\downarrow$& mIoU$\uparrow$ & \scriptsize{GFLOPs}$\downarrow$ & \scriptsize{FPS$\uparrow$} & & mIoU$\uparrow$ & \scriptsize{GFLOPs}$\downarrow$ & \scriptsize{FPS$\uparrow$}\\		
%			\midrule[0.5pt]
%			Teacher   &  \scriptsize{ResNet101} & 65.71 & \textbf{79.07} & 53.11 & 70.8 && \textbf{72.03} & 119.42 & 28.3 \\
%			Auxiliary &  ResNet50  & 46.72 & 66.38 & 33.61 & 103.5 && 62.57 & 75.55 & 42.6 \\
%			Student   &  ResNet34  & \textbf{26.28} & 57.41 & \textbf{20.43} & \textbf{230.8} && 56.86 & \textbf{45.97} & \textbf{94.1} \\
%			\toprule[0.75pt]
%	\end{tabular}
%}
%\end{table}

% -----------------  Ablation Study ------------

\subsection{Ablation Study}
To delve deeper into the individual components of the proposed method, we conduct extensive ablation studies on both the lexically-constrained pseudo-labeling module and the keyword-refined captioning module. We train the model using a three GT sentence along with three pseudo-labeled sentences. Unless otherwise stated, all settings adhere to the default configuration during training.

% ------------------ Ablation on the components under full supervision  ----------------
\begin{table*}[!t]
	\centering
	\caption{Ablation on the components of the proposed method with full supervision.}
	\label{table:abl_component_full}
	\resizebox{\textwidth}{!}{
	\setlength{\tabcolsep}{1mm}{ 
		\begin{tabular}{cc c cc c cccc c cccc c cccc}
			\toprule[0.75pt]			
			\multirow{2}{*}{OJT} & \multirow{2}{*}{KR} & \multirow{2}{*}{ } & \multicolumn{2}{c}{Sentence Decoder} & \multirow{2}{*}{ } & \multicolumn{5}{c}{MSVD\cite{chen-acl2011-msvd}} & \multicolumn{5}{c}{MSR-VTT\cite{xu-cvpr2016-msr-vtt}} & \multicolumn{4}{c}{VATEX\cite{wang-iccv2019-vatex}}  \\ 
			\cmidrule[0.5pt]{4-5} \cmidrule[0.5pt]{7-10} \cmidrule[0.5pt]{12-15} \cmidrule[0.5pt]{17-20}
			&  &  & vanilla & GF & & B@4$\uparrow$ & M$\uparrow$ & R$\uparrow$ & C$\uparrow$ & & B@4$\uparrow$ & M$\uparrow$ & R$\uparrow$ & C$\uparrow$ & & B@4$\uparrow$ & M$\uparrow$ & R$\uparrow$ & C$\uparrow$  \\ 
			\midrule[0.5pt]
			&  &  & \checkmark   &            & &56.4 &35.2 &73.2 &102.2 & &40.5 &26.6 &59.5 &48.3 & &32.4 &20.1 &46.1 &50.3 \\
			\checkmark &    &    & \checkmark  &  & &57.3 &36.1 &73.9 &104.2 & &41.8 &27.3 &60.4 &49.2 & &32.9 &20.6 &46.9 &51.2 \\
			\checkmark &\checkmark & &&& &59.6 &38.2 &75.4 &106.3 & &43.3 &29.2 &62.3 &52.4 & &33.7 &22.6 &48.3 &52.8  \\   
			\checkmark &\checkmark & &&\checkmark & &\textbf{60.1} &\textbf{39.3} &\textbf{76.2} &\textbf{107.2} 
			& &\textbf{44.8} &\textbf{30.5} &\textbf{63.4} &\textbf{53.5}
			& &\textbf{34.3} &\textbf{23.5} &\textbf{49.5} &\textbf{53.3}  \\   
			\toprule[0.75pt]
		\end{tabular}
	}
}
\end{table*}

% ------------------ Ablation on the components under few supervision  ----------------
\begin{table*}[!t]
	\centering
	\caption{Ablation on the components of the proposed method with few supervision.}
	\label{table:abl_component_few}
	\resizebox{\textwidth}{!}{
	\setlength{\tabcolsep}{0.8mm}{ 
		\begin{tabular}{cc c cc c cccc c cccc c cccc}
			\toprule[0.75pt]			
			\multirow{2}{*}{OJT} & \multirow{2}{*}{KR} & \multirow{2}{*}{ } & \multicolumn{2}{c}{Sentence Decoder} & \multirow{2}{*}{$\mathcal{L}_{word}$} & \multicolumn{5}{c}{MSVD\cite{chen-acl2011-msvd}} & \multicolumn{5}{c}{MSR-VTT\cite{xu-cvpr2016-msr-vtt}} & \multicolumn{4}{c}{VATEX\cite{wang-iccv2019-vatex}}  \\ 
			\cmidrule[0.5pt]{4-5} \cmidrule[0.5pt]{7-10} \cmidrule[0.5pt]{12-15} \cmidrule[0.5pt]{17-20}
			&  &  & vanilla & GF & & B@4$\uparrow$ & M$\uparrow$ & R$\uparrow$ & C$\uparrow$ & & B@4$\uparrow$ & M$\uparrow$ & R$\uparrow$ & C$\uparrow$ & & B@4$\uparrow$ & M$\uparrow$ & R$\uparrow$ & C$\uparrow$  \\ 
			\midrule[0.5pt]
			&  &       &\checkmark   &           & &39.5 &29.8 &64.5 & 71.2 & &32.1 &20.3 &52.1 &36.4 & &26.5 &15.1 &41.2 &38.7 \\
			\checkmark &    &    & \checkmark  & & &40.3 &30.5 &65.2 & 71.8 & &33.2 &21.4 &52.9 &37.3 & &27.6 &15.8 &42.1 &40.4 \\
			\checkmark &\checkmark & &&& &41.6 &31.5 &66.6 & 72.8 & &34.8 &22.7 &54.2 &40.8 & &29.6 &17.4 &43.2 &42.5 \\
			\checkmark &\checkmark & &&\checkmark& &42.9 &32.1 &67.8 & 73.7 & &35.1 &23.9 &55.4 &41.5 & &30.6 &18.5 &45.2 &43.7 \\
			\checkmark &\checkmark & &&\checkmark&\checkmark &\textbf{43.8} &\textbf{32.9} &\textbf{68.0} &\textbf{74.9} 
			& &\textbf{35.9} &\textbf{24.6} &\textbf{55.6} &\textbf{42.3}
			& &\textbf{31.5} &\textbf{19.4} &\textbf{46.3} &\textbf{45.2}  \\   
			\toprule[0.75pt]
		\end{tabular}
	}
}
\end{table*}

% ------------------ Ablation on the pseudo-labeling strategy ----------------
\begin{table*}[!t]
	\centering
	\caption{Ablation on the pseudo-labeling strategy.}
	\label{table:abl_pseudolabeling}
	\resizebox{\textwidth}{!}{
	\setlength{\tabcolsep}{0.6mm}{ 
		\begin{tabular}{lc cc c cccc c cccc c cccc}
			\toprule[0.75pt]
			\multirow{2}{*}{Method} & \multirow{2}{*}{Venue} & \multicolumn{2}{c}{Sentence} & \multirow{2}{*}{} & \multicolumn{5}{c}{MSVD\cite{chen-acl2011-msvd}} & \multicolumn{5}{c}{MSR-VTT\cite{xu-cvpr2016-msr-vtt}} & \multicolumn{4}{c}{VATEX\cite{wang-iccv2019-vatex}}  \\ 
			\cmidrule[0.5pt]{3-4} \cmidrule[0.5pt]{6-9} \cmidrule[0.5pt]{11-14} \cmidrule[0.5pt]{16-19}
			& & GT & Pse &  & B@4$\uparrow$ & M$\uparrow$ & R$\uparrow$ & C$\uparrow$ & & B@4$\uparrow$ & M$\uparrow$ & R$\uparrow$ & C$\uparrow$ & & B@4$\uparrow$ & M$\uparrow$ & R$\uparrow$ & C$\uparrow$  \\ 
			\midrule[0.5pt]
			Synonym Replacement\cite{wei-emnlp2019-eda}  &EMNLP'19  & 1 & 2 & &\underline{29.5}    &24.5 &61.0 &52.6 & 
			&\underline{26.1}    &18.6 &50.7 &26.4  & 
			&25.2                &16.6 &42.4 &35.2  \\ 
			Random Insertion\cite{wei-emnlp2019-eda}     &EMNLP'19 & 1 & 2 & &22.4                &22.2 &58.1 &47.8 & 
			&24.2                &17.6 &47.7 &23.3  &
			&21.9                &15.3 &42.5 &33.4  \\ 
			Back-Translation\cite{lowell-emnlp2021-masklan}      &EMNLP'21 & 1 & 2 & &18.1 &16.0  &46.0 &42.5 & 
			&17.7                &15.2                &39.3                &18.3   &
			&13.4                &11.2                &36.3                &17.3  \\  
			Masked Language Model\cite{lowell-emnlp2021-masklan} &EMNLP'21 & 1 & 2 & &26.5 &\underline{27.5}    &\underline{62.1}    &55.0   &&23.9  &20.0    &\underline{51.5}    &\underline{28.2}  &
			&\underline{27.1}    &\underline{17.1}    &\underline{43.7}    &39.3    \\			
			\toprule[0.5pt]
			Mistral-7B\cite{jiang-arxiv2023-mistral} & arXiv'23 & 1 & 2 & & 27.3 & 23.8 & 48.6 & \underline{56.4} & & 23.7 & \underline{20.2} & 46.8 & 22.6 & & 25.7 & 16.4 & 42.2 & \underline{39.8} \\
			\toprule[0.5pt]
			Ours &  & 1 & 2 & &\textbf{31.6}  &\textbf{31.3}  &\textbf{64.9} &\textbf{60.1} & 
			&\textbf{28.8}  &\textbf{21.3}  &\textbf{52.3}  &\textbf{31.4}  & 
			&\textbf{28.8}  &\textbf{18.9}  &\textbf{44.3}  &\textbf{41.2} \\
			&  &   &   & &+2.1 &+3.8  &+2.8  &+3.7 & &+2.7 &+0.9 &+0.8 &+3.2 & &+1.7  &+1.8   &+1.6 &+1.4 \\
			\toprule[0.75pt]
		\end{tabular}
	}
}
\end{table*}

\textbf{Individual Component}. We examine the components of our method, including Object-Joint Transformer (OJT), KR (Keyword Refiner), and Gated Fusion (GF), in both fully-supervised (six GT sentences) and few-supervised setting. Their results are respectively shown in Table~\ref{table:abl_component_full} and Table~\ref{table:abl_component_few}, where the top two rows employ a vanilla decoder that consists of one feed-forward network layer followed by layer normalization. This vanilla decoder serves as a contrast to our designed decoder that incorporates gated fusion. The baseline in Row~1 adopts direct concatenation of object feature and appearance-motion feature along the channel dimension without keyword refiner. In Row~2, when we add the object transformer and joint transformer, the performance is upgraded by a gain of 2.0\%/0.6\% (full/few supervision), 0.9\%/0.9\%, and 0.9\%/0.7\% CIDEr-D scores on MSVD\cite{chen-acl2011-msvd}, MSR-VTT\cite{xu-cvpr2016-msr-vtt}, and VATEX\cite{wang-iccv2019-vatex}, respectively. This might be the reason that attention-based encoder well captures the relations of object features and appearance-motion features. In Row~3, we add the keyword refiner which aligns the keywords to the video content and uses the designed decoder without gated fusion, which further promotes the performance by 2.3\%/1.3\%(full/few supervision), 1.5\%/1.3\%, and 0.8\%/1.0\% BLUE@4 scores on the three benchmarks respectively. In Row~4, we add the gated fusion strategy that considers the different contributions of visual cues and textural semantics in the decoder, which enhances the performance by 1.1\%/0.6\%(full/few supervision), 1.3\%/1.2\%, 0.9\%/1.1\% METOR scores on the three benchmarks respectively. In Row~5 of Table~\ref{table:abl_component_few}, we add the keyword loss $\mathcal{L}_{word}$ to approach the semantic similarity between pseudo-labeled and ground-truth keywords, which makes a performance rise by 1.2\%, 0.8\%, and 1.5\% CIDEr-D scores on the three benchmarks respectively. 

\textbf{Pseudo-labeling Strategy}. We compare four common sentence generation (\ie, text augmentation) methods including synonym replacement\cite{wei-emnlp2019-eda}, random insertion\cite{wei-emnlp2019-eda}, back-translation\cite{lowell-emnlp2021-masklan} , and Masked Language Model (MLM) \cite{lowell-emnlp2021-masklan}, with our lexically-constrained augmentation in Table~\ref{table:abl_pseudolabeling}. Among them, the former two methods and MLM \cite{lowell-emnlp2021-masklan} randomly select words for editing, and MLM leverages the pretrained BERT \cite{devlin-naacl-hlt2019-bert} language representation model to predict words that replace masked tokens. In contrast, the latter back-translation translates sentences into Chinese and then back to English. As shown in the table, our lexically-constrained pseudo-labeling strategy achieves the best, which is because we adopt the language model XLNet\cite{yang-nips2019-xlnet} to predict the masked token and train a token-level classifier that chooses the most possible action to be made from four choices, \ie, copy, replace, insert, and delete. This helps to generate higher-quality pseudo-labeled sentences. In addition, we adopt the large language model Mistral-7B (7–billion parameters) to generate the pseudo-labels, and the results are inferior to ours. This might be the reason that the sentences generated by Mistral-7B are not tailored for our video captioning model.

\begin{figure}[!t]
	\centering
	\subfigure[MSVD\cite{chen-acl2011-msvd}]{\includegraphics[width=0.32\linewidth]{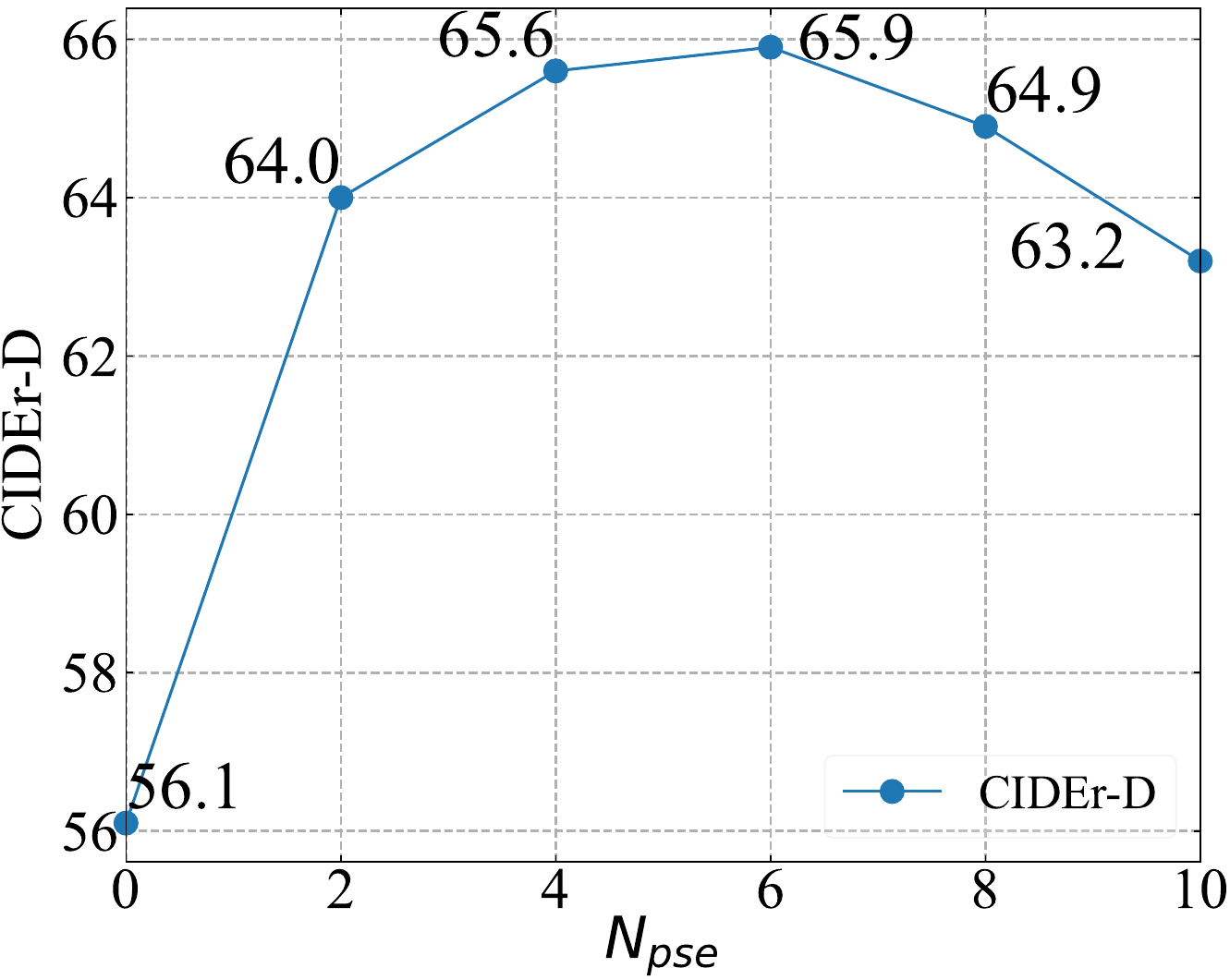}}
	\subfigure[MSR-VTT\cite{xu-cvpr2016-msr-vtt}]{\includegraphics[width=0.32\linewidth]{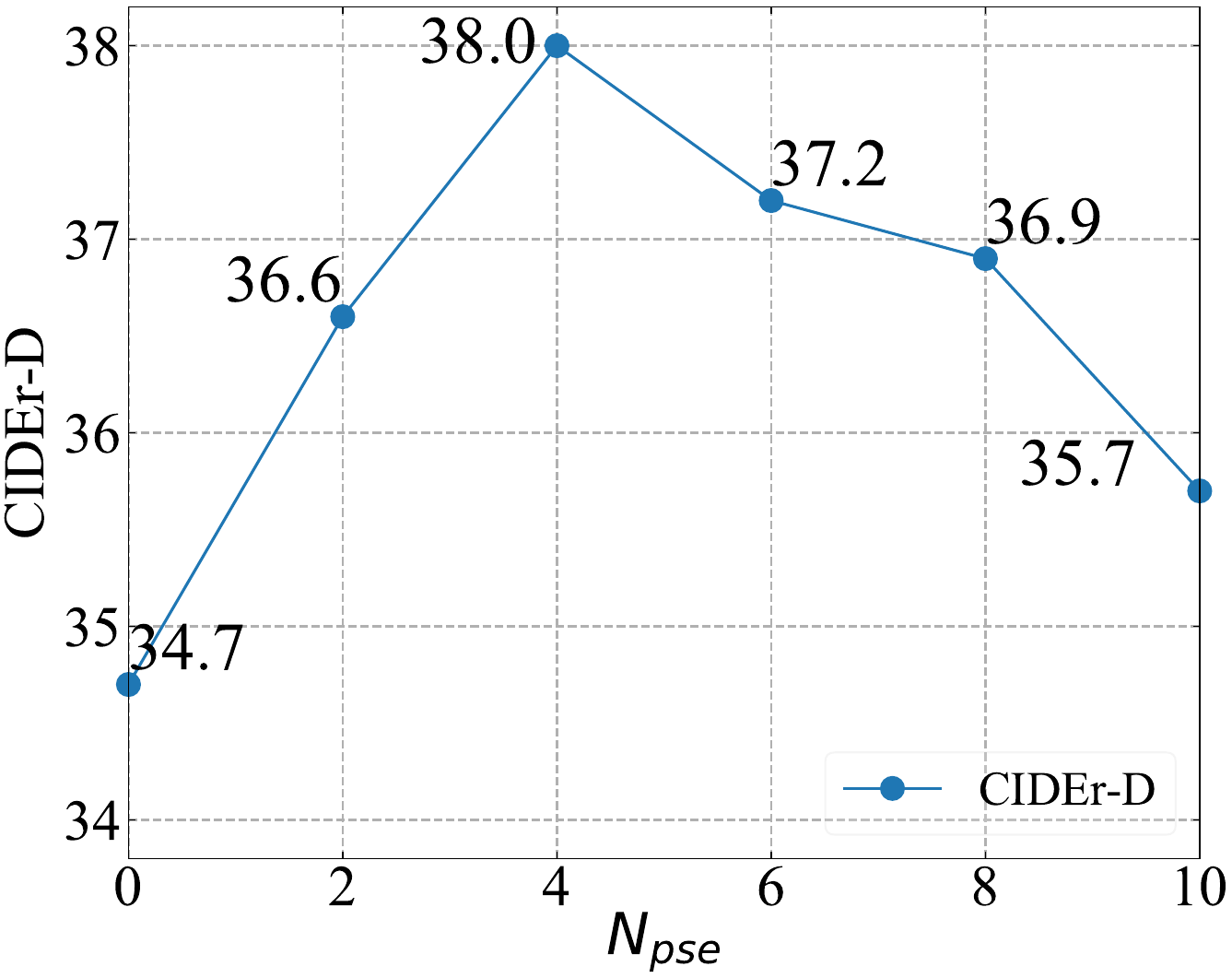}}
	\subfigure[VATEX\cite{wang-iccv2019-vatex}]{\includegraphics[width=0.32\linewidth]{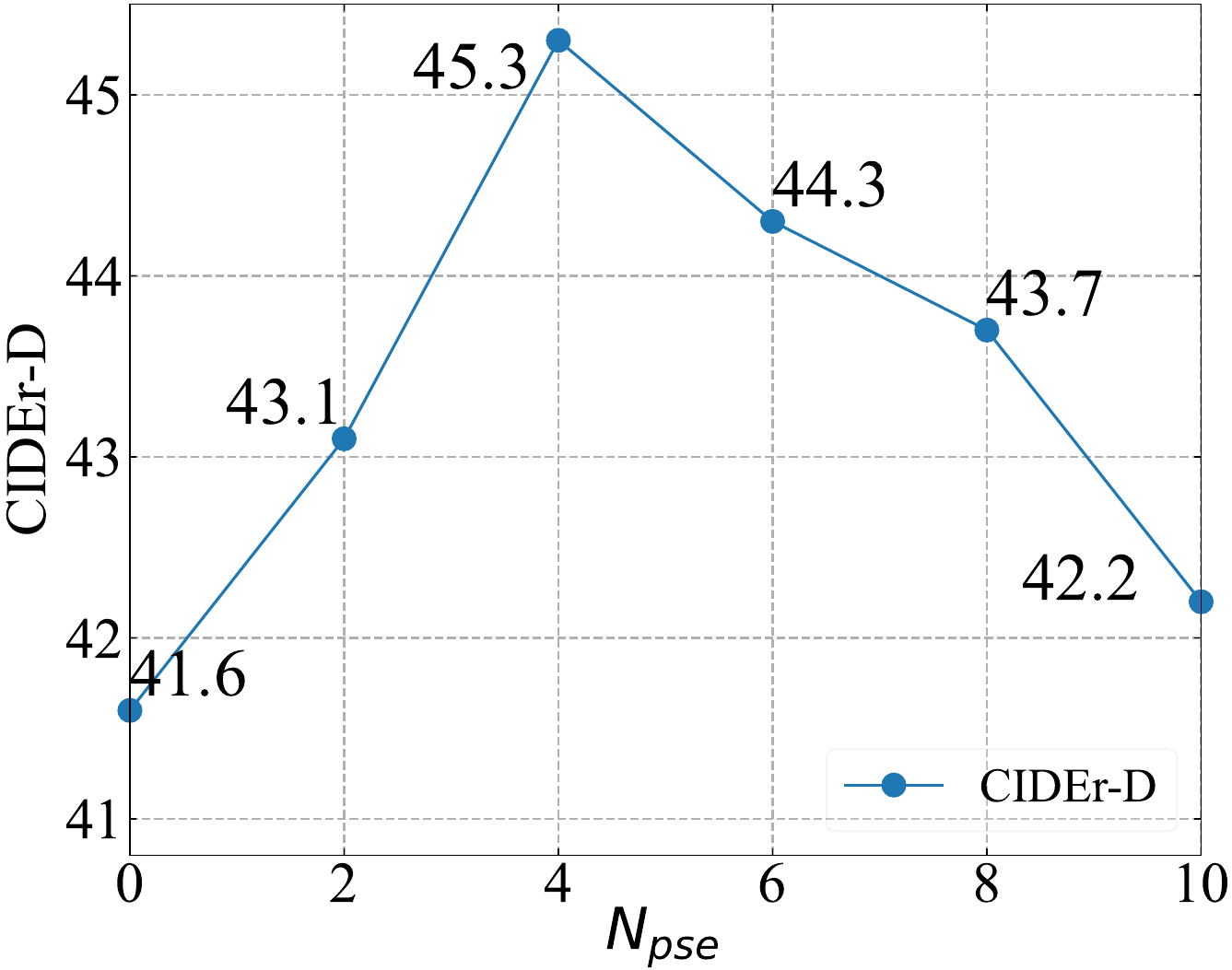}}
	\caption{Performance under different $N_{pse}$. Please zoom in for best view.}
	\label{fig:N_pse}
	%\vspace{-1mm}
\end{figure}

\textbf{Number of Pseudo-labeled Sentences $N_{pse}$}. We vary the number $N_{pse}$ of pseudo-labeled sentences from 0 to 10 at an interval of 2 and make the records in Table~\ref{table:abl_num_pseudolabel} and Fig.~\ref{fig:N_pse}. From the results, we observe that the performance saturation number is 6 on MSVD\cite{chen-acl2011-msvd}, and 4 on MSR-VTT\cite{xu-cvpr2016-msr-vtt} as well as VATEX\cite{wang-iccv2019-vatex}. In another word, the caption quality tends to weaken when the number of pseudo-labeled sentences continues to increase after the saturation point. 

% ------------- Visualization on N_pse -------

% ------------------ Ablation on N_pse number of pseudo-labels  ----------------
\begin{table*}[!t]
	\centering
	\caption{Ablation on the number of pseudo-labeled sentences $N_{pse}$.}
	\label{table:abl_num_pseudolabel}
	\setlength{\tabcolsep}{1.7mm}{ 
		\begin{tabular}{cc c cccc c cccc c cccc}
			\toprule[0.75pt]
			\multicolumn{2}{c}{Sentence} & \multirow{2}{*}{}  & \multicolumn{5}{c}{MSVD\cite{chen-acl2011-msvd}} & \multicolumn{5}{c}{MSR-VTT\cite{xu-cvpr2016-msr-vtt}} & \multicolumn{4}{c}{VATEX\cite{wang-iccv2019-vatex}}  \\ 
			\cmidrule[0.5pt]{1-2} \cmidrule[0.5pt]{4-7} \cmidrule[0.5pt]{9-12} \cmidrule[0.5pt]{14-17}
			GT & Pse &  & B@4$\uparrow$ & M$\uparrow$ & R$\uparrow$ & C$\uparrow$ & & B@4$\uparrow$ & M$\uparrow$ & R$\uparrow$ & C$\uparrow$ & & B@4$\uparrow$ & M$\uparrow$ & R$\uparrow$ & C$\uparrow$  \\ 
			\midrule[0.5pt]
			2 & 0 & &32.1          &31.7          &65.5          &56.1          & &29.0          &22.1          &52.6          &34.7  
			& &28.1          &17.9          &43.5          &41.6 \\
			2 & 2 & &41.5          &32.1          &66.9          &64.0          & &30.1          &23.1          &54.2          &36.6  
			& &29.0          &18.9          &44.8          &43.1 \\
			2 & 4 & &42.7          &32.5          &67.4          &65.6          & &\textbf{31.2} &\textbf{24.7} &\textbf{54.5} &\textbf{38.0}
			& &\textbf{29.7} &\textbf{19.2} &\textbf{45.1} &\textbf{45.3} \\
			2 & 6 & &\textbf{42.9} &\textbf{32.7} &\textbf{67.8} &\textbf{65.9} & &30.5          &23.4          &54.1          &37.2      
			& &29.2          &18.7          &44.4          &44.3 \\
			2 & 8 & &42.4          &32.2          &67.2          &64.9          & &30.1          &23.1          &53.5          &36.9 
			& &28.7          &18.5          &44.1          &43.7 \\
			2 &10 & &41.2          &31.4          &66.1          &63.2          & &29.6          &22.5          &53.1          &35.7 
			& &28.4          &18.2          &43.8          &42.2 \\
			\toprule[0.75pt]
		\end{tabular}
	}
\end{table*}

% ------------------ Ablation on L, L', L'' the number of transformer blocks ----------------
\begin{table*}[!t]
	\centering
	\caption{Ablation on the number of transformer blocks $(L/ L'/ L'')$.}
	\label{table:abl_num_transformer}
	\setlength{\tabcolsep}{0.3mm}{ 
		\resizebox{\textwidth}{!}{
			\begin{tabular}{cc cccc c cccc c cccc}
				\toprule[0.75pt]
				\multirow{2}{*}{\#} & \multirow{2}{*}{}  & \multicolumn{5}{c}{MSVD\cite{chen-acl2011-msvd}} & \multicolumn{5}{c}{MSR-VTT\cite{xu-cvpr2016-msr-vtt}} & \multicolumn{4}{c}{VATEX\cite{wang-iccv2019-vatex}}  \\ 
				\cmidrule[0.5pt]{3-6} \cmidrule[0.5pt]{8-11} \cmidrule[0.5pt]{13-16}
				&   & B@4$\uparrow$ & M$\uparrow$ & R$\uparrow$ & C$\uparrow$ & & B@4$\uparrow$ & M$\uparrow$ & R$\uparrow$ & C$\uparrow$ & & B@4$\uparrow$ & M$\uparrow$ & R$\uparrow$ & C$\uparrow$  \\ 
				\midrule[0.5pt]
				1/1/2 & &\textbf{43.8}/42.7/31.1 &\textbf{32.9}/31.5/31.1 &\textbf{68.0}/67.1/31.1 &\textbf{74.9}/73.9/31.1
				& &34.6/34.6/31.1        &23.2/22.7/31.1          &54.2/54.4/31.1          &40.6/40.8/31.1
				& &31.1/30.6/31.1 		 &19.0/18.2/31.1  		  &45.3/45.1/31.1 	       &44.2/43.4/31.1    \\  
				2/2/4 & &42.4/\textbf{43.8}/\textbf{43.8} &32.2/\textbf{32.9}/\textbf{32.9} &67.2/\textbf{68.0}/\textbf{68.0} 			 &74.2/\textbf{74.9}/\textbf{74.9}
				& &35.2/35.2/35.2        &23.8/23.9/24.1          &55.1/54.9/55.1          &41.5/41.4/41.7 
				& &\textbf{31.5}/31.1/30.6 &\textbf{19.4}/18.7/18.4 &\textbf{46.3}/45.7/45.2 &\textbf{45.2}/44.2/44.4 \\
				3/3/6 & &41.8/43.4/43.2  &31.7/32.2/32.4          &66.8/67.6/67.3          &73.6/74.4/74.1            
				& &\textbf{35.9}/\textbf{35.9}/\textbf{35.9} &\textbf{24.6}/\textbf{24.6}/\textbf{24.6} &\textbf{55.6}/\textbf{55.6}/\textbf{55.6} &\textbf{42.3}/\textbf{42.3}/\textbf{42.3}
				& &31.2/\textbf{31.5}/\textbf{31.5} &18.8/\textbf{19.4}/\textbf{19.4} &44.0/\textbf{46.3}/\textbf{46.3} &43.4/\textbf{45.2}/\textbf{45.2}   \\        
				4/4/8 & &41.5/42.2/42.6  &31.3/31.8/31.6          &65.2/66.4/66.2          &72.8/73.6/72.3
				& &35.1/35.4/35.4        &23.1/24.1/23.7          &54.6/55.3/55.6          &41.1/41.7/41.3 
				& &31.1/30.7/30.3        &18.7/18.5/18.6          &43.7/45.3/45.6          &42.4/43.8/44.6    \\          
				5/5/10 & &40.2/41.8/41.7   &30.7/31.1/31.1          &64.6/65.8/65.3          &72.1/72.7/71.4
				& &34.3/34.9/34.8          &22.7/23.5/22.8          &53.7/54.7/54.3          &40.3/40.9/40.5
				& &30.9/30.1/29.8          &17.9/17.5/17.2          &43.2/44.6/44.8          &41.6/42.6/43.7   \\         
				\toprule[0.75pt]
			\end{tabular}
	} }
\end{table*}

\textbf{Number of Transformer blocks $\{L, L', L''\}$}. We vary the number of transformer blocks in video encoder ($L$), keyword refiner ($L'$), and sentence decoder ($L''$) one by one, and show the results in Table~\ref{table:abl_num_transformer}. From the table, we see that the number of transformer blocks in sentence decoder is always more than that in video encoder and keyword refiner to achieve the best performance, \eg, $L''$=\{4, 6, 6\} is twice or triple larger than $L'$=\{2,3,3\} and $L$=\{1,3,2\} on MSVD\cite{chen-acl2011-msvd}, MSR-VTT\cite{xu-cvpr2016-msr-vtt}, and VATEX\cite{wang-iccv2019-vatex}, respectively. This indicates that it requires the strengthened attention mechanism to decode video-keyword features into sentence, in comparison to handling source object features, appearance-motion features, and keyword features. Furthermore, it is observed that when the number of transformer blocks exceeds a certain threshold, the performance tends to decline. The reason for this phenomenon might because that the model becomes more complex with more transformer blocks during training, which could contribute to over-fitting issues.

% -------------------
\begin{table*}[!t]
	\centering
	\caption{Ablations on the number of ground-truth sentences without pseudo labels.}
	\label{table:varyingGT}
	\resizebox{\textwidth}{!}{
		\setlength{\tabcolsep}{1.4mm}{
			\begin{tabular}{lccccccccccccccc}
			\toprule[0.75pt]
			\multirow{2}{*}{Models} & \multirow{2}{*}{GT} & \multicolumn{4}{c}{MSVD\cite{chen-acl2011-msvd}} & \multirow{2}{*}{GT} & \multicolumn{4}{c}{MSR-VTT\cite{xu-cvpr2016-msr-vtt}} & \multirow{2}{*}{GT} & \multicolumn{4}{c}{VATEX\cite{wang-iccv2019-vatex}}\\
			\cmidrule[0.5pt]{3-4} \cmidrule[0.5pt]{5-6} \cmidrule[0.5pt]{8-11} \cmidrule[0.5pt]{13-16} 
			& & B@4$\uparrow$ & M$\uparrow$ & R$\uparrow$ & C$\uparrow$ & & B@4$\uparrow$ & M$\uparrow$ & R$\uparrow$ & C$\uparrow$ & & B@4$\uparrow$ & M$\uparrow$ & R$\uparrow$ & C$\uparrow$\\
			\midrule[0.5pt]
			HMN\cite{ye-cvpr2022-hmn}  & 1  & 18.9  & 24.2  & 60.7 & 19.3 & 1 & 22.3 & 17.3 & 49.2 & 23.2 & 1 & 24.6 & 15.7 & 41.5 & 32.1 \\
			Ours & 1  & 23.3  & 27.1  & 63.5  & 21.7 & 1 & 26.2 & 18.8 & 50.3 & 26.4 & 1 & 26.9 & 17.1 & 43.6 & 36.8\\
			\midrule[0.5pt]
			HMN\cite{ye-cvpr2022-hmn} & 5  & 42.1  & 30.5  & 65.3  & 72.8 & 3 & 30.3 & 22.3 & 52.4 & 32.9 & 2 & 26.3 & 16.5 & 42.8 & 38.2 \\
			Ours & 5  & 44.4  & 33.6  & 67.9  & 75.5 & 3 & 34.6 & 24.3 & 54.9 & 37.9 & 2 & 29.4 & 18.7 & 44.8 & 41.5\\
			\midrule[0.5pt]
			HMN\cite{ye-cvpr2022-hmn} & 10 & 47.4  & 35.3  & 71.8  & 78.3 & 6 & 36.5 & 24.8 & 54.9 & 41.6 & 4 & 29.2 & 17.9 & 44.6 & 42.0\\
			Ours & 10  & 47.1  & 34.3  & 71.7  & 95.2 & 6 & 37.8 & 25.4 & 55.6 & 44.7 & 4 & 32.1 & 20.8 & 47.1 & 47.9\\
			\midrule[0.5pt]
			HMN\cite{ye-cvpr2022-hmn} & 15  & 51.1  & 36.2  & 73.2  & 87.2 & 9 & 41.3 & 25.7 & 58.1 & 43.8 & 6 & 31.2 & 18.7 & 46.1 & 44.1 \\
			Ours & 15  & 52.6  & 36.7  & 73.5  & 101.4 & 9 & 39.1 & 26.7 & 57.1 & 48.8 & 6 & 32.5 & 21.4 & 48.1 & 50.4\\ 
			\midrule[0.5pt]
			HMN\cite{ye-cvpr2022-hmn} & 20  & 54.8  & 36.7  & 74.1  & 96.5 & 12 & 42.5 & 26.6 & 59.2 & 46.3 & 8 & 31.9 & 19.3 & 47.4 & 46.5 \\
			Ours & 20  & 55.7  & 37.3  & 75.2  & 104.3 & 12 & 41.9 & 27.5 & 58.4 & 51.2 & 8 & 33.6 & 22.8 & 48.9 & 52.7\\
			\midrule[0.5pt]
			HMN\cite{ye-cvpr2022-hmn} & 30  & 57.9  & 37.2  & 74.6  & 101.3 & 15 & 43.1 & 27.2 & 59.8 & 48.9 & 10 & 32.7 & 19.8 & 48.1 & 47.9 \\
			Ours & 30  & 58.4  & 38.7  & 75.7  & 106.5 & 15 & 42.5 & 28.4 & 62.5 & 52.1 & 10 & 34.3 & 23.5 & 49.5 & 53.3\\
			\midrule[0.5pt]
			HMN\cite{ye-cvpr2022-hmn} & 40  & 59.2  & 37.7  & 75.1  & 104.0 & 20 & 43.5 & 29.0 & 62.7 & 51.5 & - & - & - & - & -\\
			Ours & 40  & 60.1  & 39.3  & 76.2  & 107.2 & 20 & 44.8 & 30.5 & 63.4 & 53.5 & - & - & - & - & -\\
			\toprule[0.75pt]
		\end{tabular}
		}
	}
\end{table*}

\textbf{Number of GT captions without pseudo labels}. To investigate the performance of the fully-supervised models with less supervision, we vary the number of Ground-Truth (GT) captions without introducing pseudo labels on the most competing HMN\cite{ye-cvpr2022-hmn} and our full-supervision version. The comparison results are shown in Table~\ref{table:varyingGT}, which indicates that the performance of the full supervised models degrade gradually when the number of GT captions reduces. Especially, the performance degrades greatly when only one caption is available, \eg, it (\eg, HMN) achieves only 18.9\% in terms of B\@4 on MSVD, which is less than one-third of that with full supervision. This suggests the necessity of the high-quality pseudo-labeling strategy under few supervision, \ie, there are only one or very few (\eg, less than 3) GT sentences.

\subsection{Qualitative Results}
\begin{figure*}[!t]
	\centering
	\subfigure[MSVD\cite{chen-acl2011-msvd}]{\label{fig:vis_aug_msvd}\includegraphics[width=0.32\linewidth]{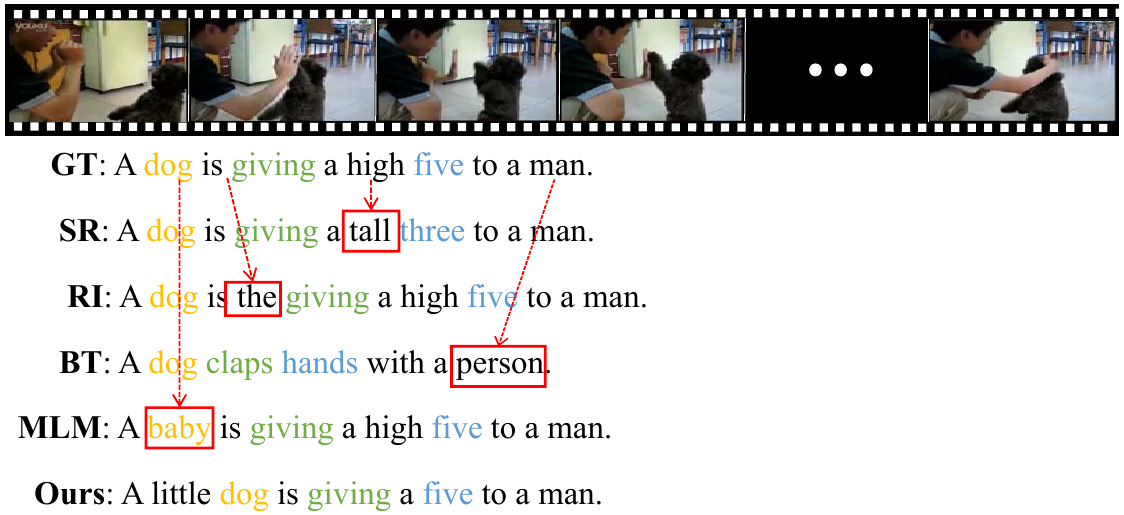}}
	\subfigure[MSR-VTT\cite{xu-cvpr2016-msr-vtt}]{\label{fig:vis_aug_msrvtt}\includegraphics[width=0.32\linewidth]{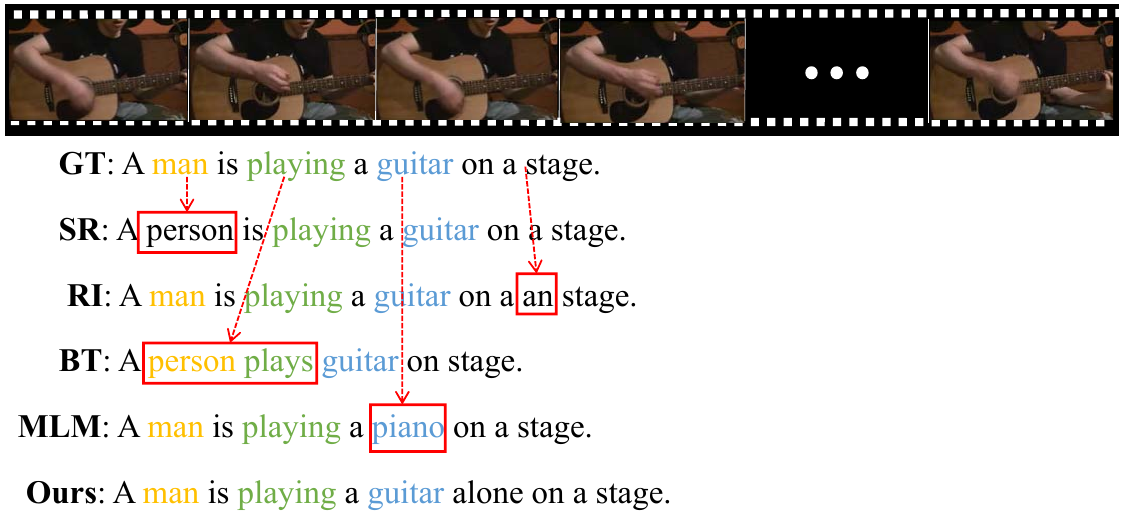}}
	\subfigure[VATEX\cite{wang-iccv2019-vatex}]{\label{fig:vis_aug_vatex}\includegraphics[width=0.32\linewidth]{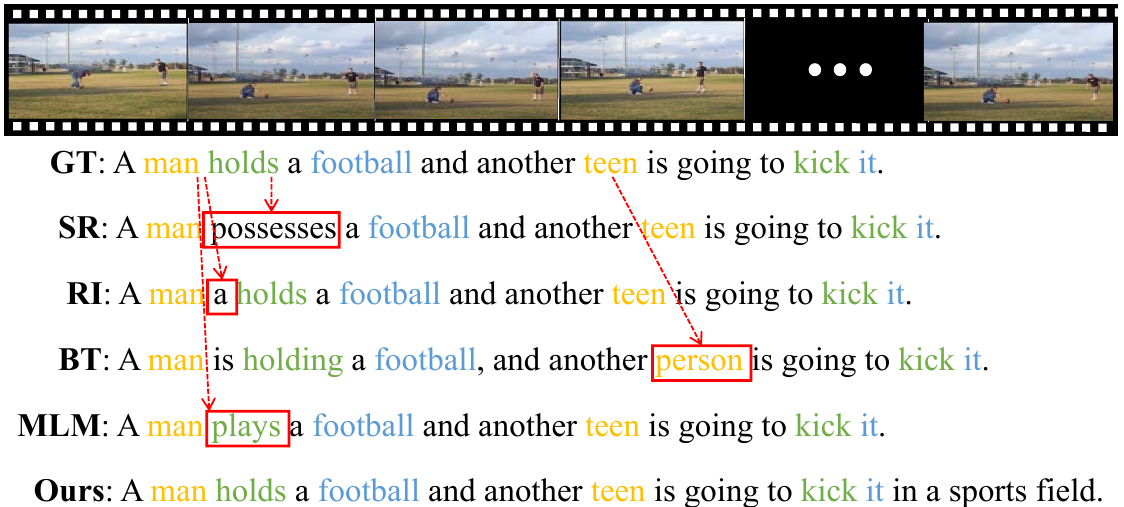}}
	\caption{Qualitative results of different pseudo-labeling strategies.}
	\label{fig:vis_augment}
	%\vspace{-1mm}
\end{figure*}

To illustrate the captions generated by adopting different pseudo-labeling strategies, we randomly select one video from test set of MSVD\cite{chen-acl2011-msvd}, MSR-VTT\cite{xu-cvpr2016-msr-vtt}, and VATEX\cite{wang-iccv2019-vatex}, respectively, whose qualitative results are depicted in Fig.~\ref{fig:vis_augment}. Here, the compared alternatives include Synonym Replacement (SR)\cite{wei-emnlp2019-eda}, Random Insertion (RI)\cite{wei-emnlp2019-eda}, Back-Translation (BT)\cite{lowell-emnlp2021-masklan}, and Masked Language Model (MLM)\cite{lowell-emnlp2021-masklan}. The subjects (\emph{yellow}), predicates (\emph{green}), and objects (\emph{blue}) in these sentences are in different color. 

From the results, it is clearly shown that our method generates higher-quality of captions compared to the rest. This is because we adopted the lexically-constrained pseudo-labeling strategy, which creates a synthetic sentence dataset by finetuning a pretrained language model that results in a token-level classifier. This classifier together with repetition penalized sampling contributes to generate more promising pseudo-labeled sentences. On the contrary, there exist some drawbacks in other methods. Among them, synonym replacement may cause semantic confusion (Row~2), \eg, replace ``high" to ``tall" in Fig.~\ref{fig:vis_aug_msvd}, ``man" to ``person" in Fig.~\ref{fig:vis_aug_msrvtt}, ``hold" to ``possess" in Fig.~\ref{fig:vis_aug_vatex}; random insertion may cause redundant words (Row~3), \eg, ``the" before ``giving" in Fig.~\ref{fig:vis_aug_msvd}, ``an" after ``a" in Fig.~\ref{fig:vis_aug_msrvtt}, ``a" before ``hold" in Fig.~\ref{fig:vis_aug_vatex}; back-translation may incur word ambiguity, \eg, ``man" to ``person" in Fig.~\ref{fig:vis_aug_msvd} and ``teen" to ``person" in Fig.~\ref{fig:vis_aug_vatex}; the masked language model may replace or insert the incorrect token, \eg, ``dog" to ``baby" in Fig.~\ref{fig:vis_aug_msvd}, ``guitar" to ``piano"  in Fig.~\ref{fig:vis_aug_msrvtt}, and ``play" before ``hold" in Fig.~\ref{fig:vis_aug_vatex}.

% ------------- Visualization on Data Augmentation -------

% ------------- Visualization of failure cases -------
\begin{figure*}[!t]
	\centering
	\subfigure[MSVD\cite{chen-acl2011-msvd}]{\label{fig:vis_fail_msvd}\includegraphics[width=0.31\linewidth]{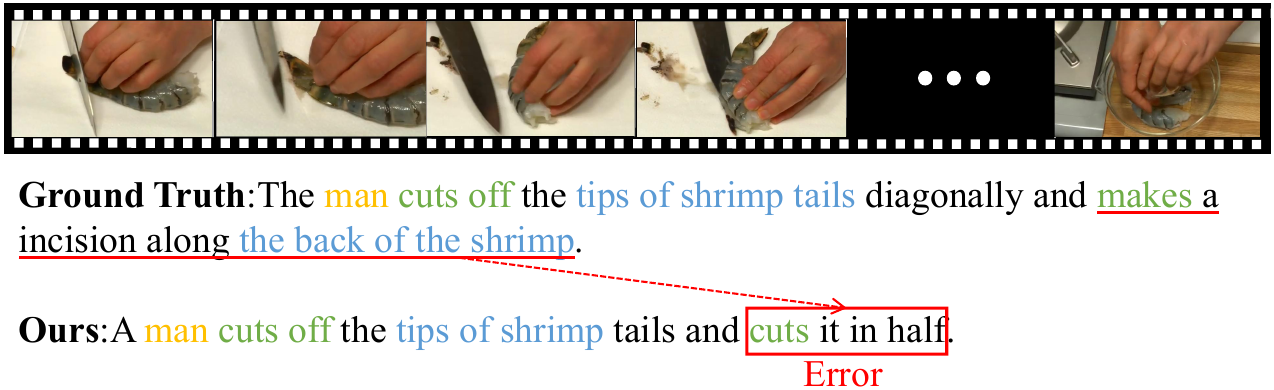}}
	\subfigure[MSR-VTT\cite{xu-cvpr2016-msr-vtt}]{\label{fig:vis_fail_msrvtt}\includegraphics[width=0.305\linewidth]{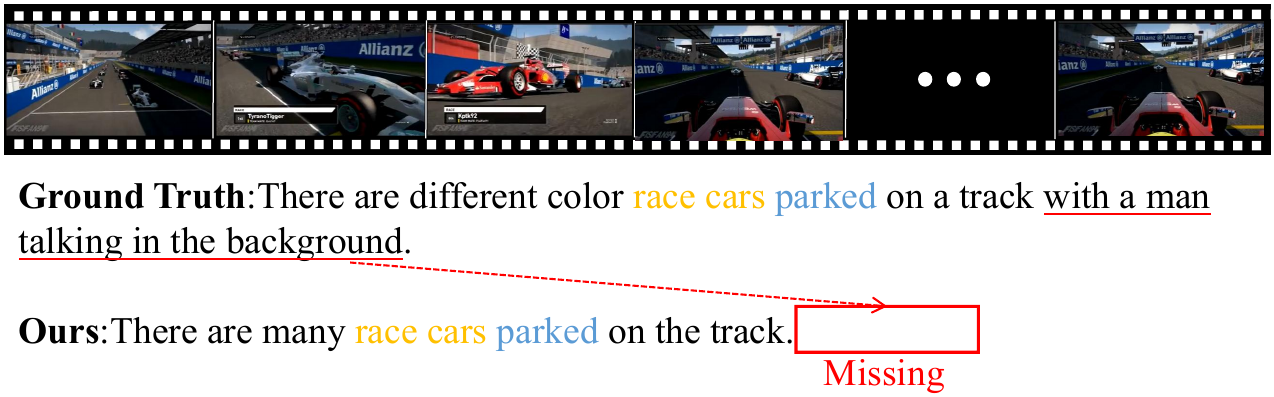}}
	\subfigure[VATEX\cite{wang-iccv2019-vatex}]{\label{fig:vis_fail_vatex}\includegraphics[width=0.34\linewidth]{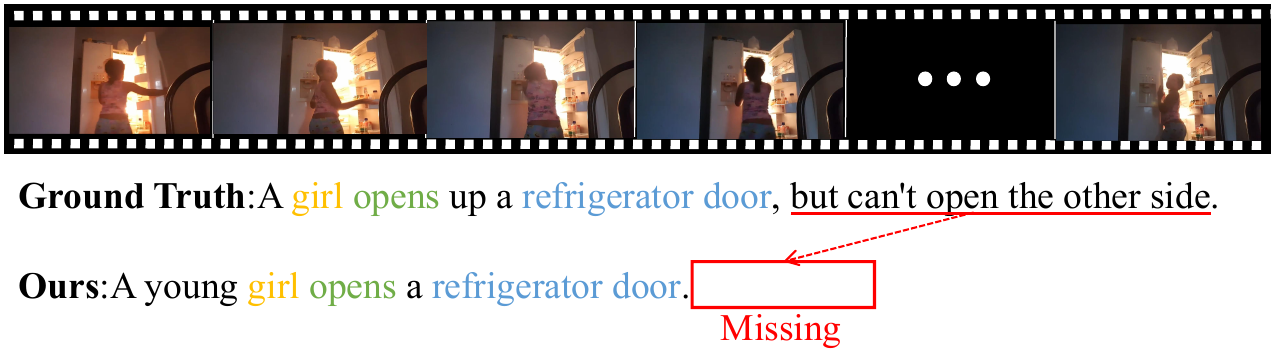}}
	\caption{{\b Qualitative results of some failure cases.}}
	\label{fig:vis_fail}
	%\vspace{-1mm}
\end{figure*}

In addition, we illustrate some failure cases caused by our model on the three data sets in Fig.~\ref{fig:vis_fail}. From these figures, we see there are some action confusion, \eg, ``makes a incision" versus ``cuts it in half" in Fig.~\ref{fig:vis_fail_msvd}, and also some actions missing, \eg, ``with a man talking in the background" in Fig.~\ref{fig:vis_fail_msrvtt} and ``but can't open the other side" in Fig.~\ref{fig:vis_fail_vatex}. These failure cases provide some insights of addressing its limitations, \eg, we may take advantage of pre-trained video multi-modal language model to generate initial captions, and refine these caption by modeling human-object interactions, which plays an important role in revealing what happens in a video.

\section{Conclusion}
\label{conclusion}
This work studies the video captioning task under few supervision, \ie, only a single or very few ground-truth sentences are available, which still remains unexplored. To provide more prior knowledge, we introduce a lexically-constrained pseudo-labeling strategy which generates pseudo-labeled sentences. This strategy creates a synthetic sentence data set by employing a pretrained language model to learn a token-level classifier, which identifies the editing action to be done on the current token. Meanwhile, we design a keyword-refined captioning module that polishes the keywords by aligning textual semantics to video content. In addition, a gated fusion scheme is incorporated to the decoder by considering the different contributions of video features and text features. Comprehensive experiments were carried out across three benchmarks demonstrate the superiority of the proposed approach. 

However, there still exist some limitations of this work. First, it still requires few ground-truth sentences to guide the generation of pseudo labels. Second, there still exists the semantic error of the objects with similar appearance. In the future, we may employ large-scale web video-text data for pre-training to improve the performance of video captioning without human-labeling sentences. Also, it is interesting to employ cross-modality data including video, audio, and text, to enhance the visual-text alignment in the embedding space.

\section*{Acknowledgement}
This work was supported in part by Zhejiang Provincial Natural Science Foundation of China under Grants LR23F020002, in part by “Pioneer” and “Leading Goose” R\&D Program of Zhejiang under Grant 2024C01075.

%National Nature Science Foundation of China (Grant No: 62273303)

\small
%\bibliography{fewcaption}

\end{document}